\documentclass{article} 
\usepackage{iclr2027_conference,times}

\usepackage{amsmath,amsfonts,amssymb,bm}

\def\eqref#1{equation~\ref{#1}}

\def\1{\bm{1}}

\DeclareMathAlphabet{\mathsfit}{\encodingdefault}{\sfdefault}{m}{sl}
\SetMathAlphabet{\mathsfit}{bold}{\encodingdefault}{\sfdefault}{bx}{n}

\usepackage{hyperref}
\usepackage{url}
\usepackage{wrapfig}
\usepackage{wasysym}
\usepackage{booktabs}
\usepackage{xspace}
\usepackage[table]{xcolor}
\usepackage{multirow}
\usepackage{caption}
\usepackage{tikz}
\usetikzlibrary{positioning, arrows.meta, calc}

\newcommand{\bayeslm}{\textsc{BayesLM}\xspace}
\newcommand{\oraclelm}{\textsc{OracleLM}\xspace}
\newcommand{\baselm}{\textsc{StartLM}\xspace}
\newcommand{\bayesianassistant}{\textsc{BayesAssist}\xspace}

\title{Bayesian Fine-tuning Yields Language Models that are as Bayesian as their Beliefs Allow}

\author{Polina Tsvilodub$^{1*}$, Andreas Waldis$^{1*}$, Linlu Qiu$^{2}$, Tal Linzen$^{3}$, Michael Franke$^{1}$\\
$^{*}$Equal contributions.\\
$^{1}$University of T\"ubingen, $^{2}$Massachusetts Institute of Technology, $^{3}$New York University\\
\texttt{polina.tsvilodub@uni-tuebingen.de}
}

\usepackage[dvipsnames]{xcolor}
\usepackage{graphicx}
\usepackage{comment}
\newcommand{\mf}[1]{\textcolor{BrickRed}{[MF: #1]}}

\newcommand{\pt}[1]{\textcolor{Cerulean}{[PT: #1]}}
\usepackage{mdframed}
\newmdenv[
  topline=false,
  bottomline=false,
  rightline=false,
  linecolor=gray,
  linewidth=1pt,
  skipabove=\topsep,
  skipbelow=\topsep,
  fontcolor=gray
]{draftytext}
\iclrfinalcopy 

\begin{document}

\maketitle

\begin{abstract}
Language models (LMs) are increasingly used for tasks that require reasoning about hidden variables from a few observations, for which Bayesian inference is the normatively correct solution.
While supervised fine-tuning of an LM on the outputs of an optimal \textit{Bayesian} model leads to near-Bayesian behavior, standard supervised fine-tuning (SFT) on the true answers for the task falls short of it.
But behavior alone does not tell us \textit{why} tuning on a \textit{Bayesian} or an \textit{oracle} (true answers) signal differs: whether the resulting LM represents Bayesian beliefs, acts on them, or turns them into a choice the way Bayes' rule does.
To compare them, we formulate increasingly demanding requirements for an LM to count as a Bayesian decision maker, spanning its behavior, representations, and computations, and test them on a flight recommendation task.
The Bayes-trained LM acts Bayesian, encodes quantities of Bayes' rule in its middle layers, and uses the encoded belief for the recommendation to a certain extent. 
The oracle-trained LM differs from it 
both in the beliefs it holds and whether it reads beliefs out into recommendations.  
Exchanging beliefs between the LMs transfers a part of the Bayesian advantage.
Bayes fine-tuning thus installs usable Bayesian beliefs in an LM for reasoning under uncertainty in a way standard SFT on oracle answers cannot, highlighting the advantage of nuanced supervision.

\end{abstract}
\hspace{14em}\includegraphics[width=2em,height=2em]{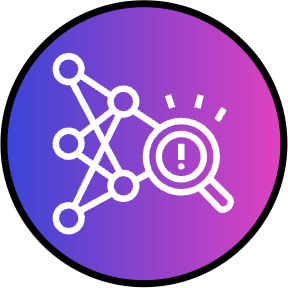}\hspace{.75em}\parbox{40em}{\href{https://bayeslm.github.io}{\vspace*{1.3em}\texttt{bayeslm.github.io}}
}\hfill\null
\vspace{-2em}

\section{Introduction}
\label{sec:introduction}

\begin{figure}[t]
  \centering
  \includegraphics[width=\textwidth]{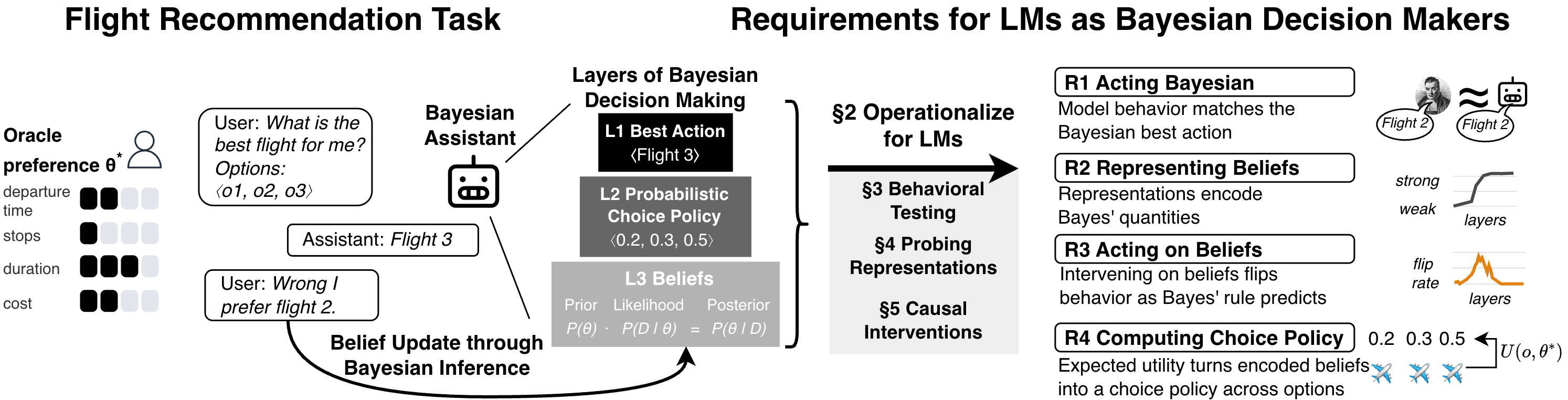}
  \caption{\textbf{Overview of our framework for analyzing to which extent LMs instantiate Bayesian inference}. We test the framework on a flight recommendation task (left), which is optimally solved through Bayesian decision making (middle). We operationalize the consistency with Bayes-optimal decision making in LMs through four increasingly fine-grained requirements capturing different layers of Bayesian inference (right), showing that LMs are less consistent with increasing requirements.}
  \label{fig:overview}
\end{figure}

Language models (LMs) excel at many tasks, including those that require reasoning under uncertainty, such as inferring a user's latent preferences to make adequate action recommendations, as in the task in focus in this paper \citep{brown2020language, shao2024deepseekmath}.
While pretrained LMs often fail on such tasks \citep{pournemat2025reasoning, samanta2026bayesbench}, and standard supervised fine-tuning on \textit{golden} task labels (SFT) only leads to suboptimal performance, specific \textit{Bayesian} supervised fine-tuning can produce LMs with behavior that matches a Bayesian decision maker \textit{superficially} \citep{qiu2026bayesian}.
However, superficial Bayes-like behavior does not guarantee following the hierarchical layering of computational steps suggested by normative Bayesian theory (see Figure~\ref{fig:overview}):
choosing an action (layer L1), based on a policy derived from the expected utility over relevant alternatives (L2), which is based on uncertain beliefs informed by the observational evidence (L3).
In principle, an LM could show Bayesian behavior (L1) without showing internal Bayesian computation at ``deeper'' levels (L2 or L3), especially since in the normative Bayesian framework the mappings from beliefs to policies, and from policies to best choices are many-to-one.
Reversely, though perhaps less likely \textit{a priori}, as each computational step presupposes the previous one, an LM may also master the earlier steps but fail to reach the correct action.
Understanding these computational steps in LMs is not only important for predictability \citep[e.g.,][]{Mitchell2026:Jagged-Intellig} but crucially understudied.

We address this gap and ask here how consistently LMs follow these layers of Bayesian decision making.
We test each layer with behavioral evaluation, probing of representations, and causal interventions, organized as four increasingly demanding requirements that, if all met, would constitute converging evidence for genuine computation of Bayes' rule in an LM (\autoref{fig:overview}, right): that the LM \textbf{acts Bayesian} (\textbf{R1}, \autoref{sec:acting-bayesian}), that it \textbf{represents} the beliefs Bayes' rule prescribes (\textbf{R2}, \autoref{sec:representing-beliefs}), that it causally \textbf{acts on those beliefs} (\textbf{R3}, \autoref{sec:acting-on-beliefs}), and that it \textbf{computes} a choice policy in a Bayes optimal way using these beliefs (\textbf{R4}, \autoref{sec:bayesian-implementation}).
We apply these tests to a triple of LMs from \citet{qiu2026bayesian}: an unmodified pretrained model (the \baselm{}), the \bayeslm{}, fine-tuned to match the flight recommendations of an ideal Bayesian decision maker (\bayesianassistant{}), and the \oraclelm{}, fine-tuned on the true flight preference (oracle).
The two fine-tuned models share the \baselm{}, the task, and the fine-tuning procedure, and differ only in the training signal.
Comparing them against R1--R4 thus lets us locate where oracle (\oraclelm{}) diverges from Bayesian fine-tuning (\bayeslm{}).
If the \bayeslm{} exhibits genuine Bayesian decision making, we expect it to be more Bayes-consistent than the \oraclelm{} not only in its behavior but also in its representations and computations.

At R1 and R2, we find that \bayeslm{}, and \oraclelm{} to a lesser degree, does match the \bayesianassistant{}: its best action (L1), its choice policy (L2), and the quantities of Bayes' rule encoded in its representations (L3) are all Bayes-consistent.
Testing R3, we find partial consistency only in \bayeslm{}.
Causal interventions let us inject a specific prior about the user's preferences, most effectively in the layers identified under R2, but no single token or layer carries the effect.
Testing R4 via plugging LMs' beliefs into the explicit assistant, we find that the step from beliefs (L3) to choice policy (L2) follows expected utility for the \bayeslm{} more than for the \oraclelm{}:
the \bayeslm{} is as Bayes-accurate as its beliefs allow,
while \oraclelm{}'s beliefs carry more information than is read-out for behavioral predictions.
Patching beliefs between the two models confirms this: the \bayeslm{}'s beliefs raise the \oraclelm{}'s Bayes accuracy, but not to the same level as \bayeslm{}. 
\textbf{Thus, Bayesian fine-tuning installs Bayes-consistent beliefs and a read-out that forms a more optimal choice policy that oracle fine-tuning does not install.}\footnote{All materials are available at \url{https://anonymous.4open.science/r/bayes-paper-3CF2}.}

\section{Experimental set-up}\label{sec:setup}

We use the flight recommendation task and the models of \citet{qiu2026bayesian}. 
This section introduces the task, the normatively correct Bayesian assistant, the language models to be investigated, as well as target measures, terminology, notation and the test sets used in the following analysis sections.

\paragraph{Recommendation task.}
The flight recommendation task of \citet{qiu2026bayesian} has two actors: the user and the assistant. 
Each trial consists of five consecutive rounds, $D_i = \langle d_1, …, d_i \rangle$ for $1 \le i \le 5$, where the single data point $d_i = \langle O_i,c_i \rangle$ for round $i$, and where the assistant recommends one of three flight options $O_i =\ \{o_1, o_2, o_3\}$ for the user in each round (\autoref{fig:overview}, left). 
Each flight option is represented as a four-place vector $o_j \in \mathbb{R}^4$ of numerical values for each of the four features (departure time, duration, number of stops and a price), 
and the user has some weighted \textit{preference} $\theta$ over these features, which are unknown to the assistant.
Following \citet{qiu2026bayesian}, we only consider preference values from the set $\{-1, -0.5, 0, 0.5, 1 \}$, with $\Theta$ as the set of all valid preference value vectors.
The assistant 
observes the user's most preferred option $c_i \in O_i$ at the end of each round $i$, which allows the assistant to sequentially update their beliefs about the user's latent preference, which is held constant.
Each $c_i$ is the outcome of (strict) maximization of utility for a fixed $\theta$.
The \textit{utility} $U(o_j, \theta)$ of option $o_j$ under preference $\theta$ is given by the vector dot-product $U(o_j, \theta) = o_j \cdot \theta = \sum_{k=1}^{4} o_{jk} \ \theta_k$.


\paragraph{Bayesian Assistant.}
Since the user's preferences are not directly observable, the optimal \bayesianassistant{} captures \textit{uncertainty} about them as a \textit{probabilistic belief about the latent preference} $P \in \Delta(\Theta)$. 
This belief is updated based on $D_i$ by computing the \textit{posterior beliefs} $P(\theta \mid D_i)$ via Bayes' rule.
To capture the round-based task structure, the Bayesian posterior can be written as a sequence of belief updates via recursion over sets $D_{\le i}$ containing the first $i$ rounds:
\begin{equation}
\label{eq:sequential-bayesian-assistant}
P(\theta \mid D_{\le i}) \propto P(c_i \mid o_i, \theta) \times P(\theta \mid D_{\le i-1}), \quad i = 1, \dots, 5,\ \text{with } P(\theta \mid D_{\le 0}) := P(\theta)
\end{equation}
Throughout, we use the notation ``$\text{belief}_i$'' to indicate the posterior beliefs after seeing $i$ data points.
So, the initial prior beliefs $P(\theta)$ before any observation of user choice is written as ``$\text{belief}_0$''.
Note that $\text{belief}_i$ is 
also the assistant's \textit{prior} for making a recommendation at round $i+1$.

Given any belief $P_{\Theta}$ about preferences and a triple of options $\langle o_1, o_2 , o_{3} \rangle$, the \textit{expected utility} of option $o_j$ is defined as $\text{EU}(o_j, P_\Theta) = \sum_\theta P_\Theta(\theta) U(o_j, \theta)$.\footnote{\citet{qiu2026bayesian} assume that each feature dimension is treated independently in $P_\Theta$, so expected utility maximization reduces to a best response to the feature-wise mean beliefs.}
We consider the \textit{Bayesian policy} for (approximately) rational choice under softmax-optimization of $\text{EU}$, with scalable rationality parameter $\beta$. 
Notice that the softmax operation entails that the ratio of probabilities assigned to any two options is a function of the difference in \textit{EU} between them, so that soft policies predict mistakes when the differences between best and second-best options are small (which are more or less pronounced depending on the softmax parameter $\beta)$.
The (strict) \textit{Bayes-optimal recommendation} is the (in our case: unique) option $o_i$ with positive probability under the Bayesian policy with $\beta \rightarrow \infty$.

\paragraph{Language models.}
We test three related LMs from \citet{qiu2026bayesian}. 
A \baselm{}, the \bayeslm{} fine-tuned on the \bayesianassistant{}'s recommendations at each round, and the \oraclelm{} fine-tuned on the oracle (the true user choice).
Comparing the latter two isolates the effect of the standard oracle signal against the Bayesian one.
We focus on Gemma-2 9B \citep{team2024gemma} and report Llama-3.1 8B \citep{grattafiori2024llama} and Qwen-2.5 7B \citep{qwen2025qwen25technicalreport} in \autoref{app:backbones}.

\paragraph{LMs' choices, policies and beliefs.}
We compare the predictions of the different LMs against the normative \bayesianassistant{} at three different layers of analysis (see \autoref{fig:overview}).
\textbf{L1} is the \textit{best action}, the option the policy recommends;
\textbf{L2} is the \textit{choice policy} that turns $P_{\Theta}$ into a probability distribution over the three flight options;
\textbf{L3} are the \textit{beliefs} $P_{\Theta}$ about the user's latent preference, given by the prior, likelihood and posterior of Bayes' rule.
For the \bayesianassistant{}, L1 to L3 are given.
For the LMs, the best action (L1) is the arg-max of their policy.
The \textit{LM policy} (L2) is the conditional probability the LM assigns to the flight option tokens (``1'', ``2'', or ``3'') after the prompt that corresponds to the data points $D_{\le i}$.
The LMs' beliefs (L3) are not transparent, but are here  obtained via linear probes and referred to as \textit{decoded beliefs}.
In doing so, we linearly probe also for LMs' representations of the \textit{decoded likelihood} (see below).
We also obtain the \textit{decoded policy} and the \textit{decoded best action} by probing the LM's choice policy.
In this way, we obtain an internal counterpart to L1 and L2 to check whether and how LMs align internally with what they emit. 

\paragraph{Measures of accuracy and consistency.}
We account for absolute goodness-of-fit via corresponding measures 
across the three layers of Bayesian decision making. 
On L1, we call an LM recommendation \textit{Bayes correct} if it matches the Bayesian recommendation, and \textit{Oracle correct} if it matches the true user choice, allowing us to calculate \textit{Bayes accuracy} or \textit{Oracle accuracy}, accordingly. 
Note that the \bayesianassistant{}'s oracle accuracy is not perfect (see \autoref{app:acting-bayesian}).
At L2, we consider the \textit{(decoded) policy divergence} (for a given $\beta$) as the KL-divergence between the \bayesianassistant{}'s policy for softmax parameter $\beta$ and the LM (decoded) policy.
For L3, the \textit{belief divergence} is the KL-divergence between the \bayesianassistant{}'s beliefs and the LM's decoded belief.

\paragraph{Test sets.} 
We evaluate these models on four test sets that isolate a single property of Bayes' rule.
The \textit{original} set, held out from fine-tuning \citep{qiu2026bayesian}, tests whether a measure matches the assistant at all, and the others additionally test for Bayes consistency.
The \textit{shuffled} set reorders the evidence within the first four rounds and tests the expected invariance to evidence order. 
On the \textit{low noise} (one user's choices is random), we expect small effects as the assistant's beliefs mostly recover by round 5; on the \textit{high noise} set (three random choices), we expect a substantial effect where the assistant remains uncertain. 
    As predicted by the \bayesianassistant{} and confirmed by comparing the \bayeslm{}'s performance on original and shuffled datasets, the models are invariant to the order of evidence. Accordingly, the location of noise injections in the noisy test sets is not a relevant factor.

\section{Acting Bayesian: Behavioral Analysis}
\label{sec:acting-bayesian}

To address the behavioral requirement \textbf{R1}, we investigate the Bayes accuracy and the policy divergence of \bayeslm (L1, L2) and compare it with \oraclelm{}, and \baselm{} as a lower bound.

\begin{figure*}[t]
  \centering
  \begin{minipage}[b]{0.55\textwidth}
    \centering
      \includegraphics[width=\linewidth]{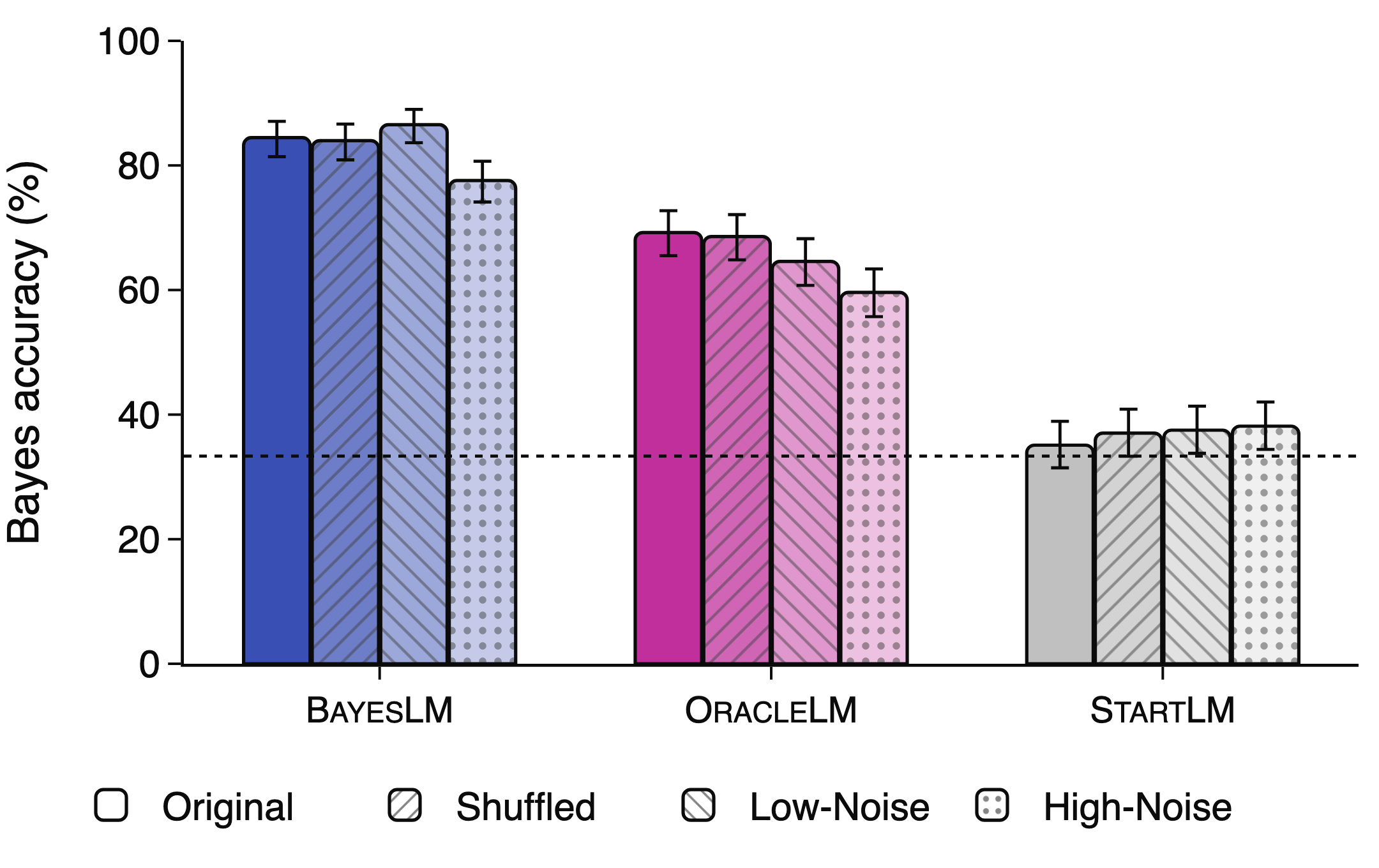}
      \caption{Bayes accuracy on the four test sets, for the \bayeslm{}, \oraclelm{}, and \baselm{}.
      Chance agreement is at $1/3$ (dashed line).}
      \label{fig:behavioral-accuracy}

  \end{minipage}
  \hfill
  \begin{minipage}[b]{0.42\textwidth}
    \centering
    \includegraphics[width=\linewidth]{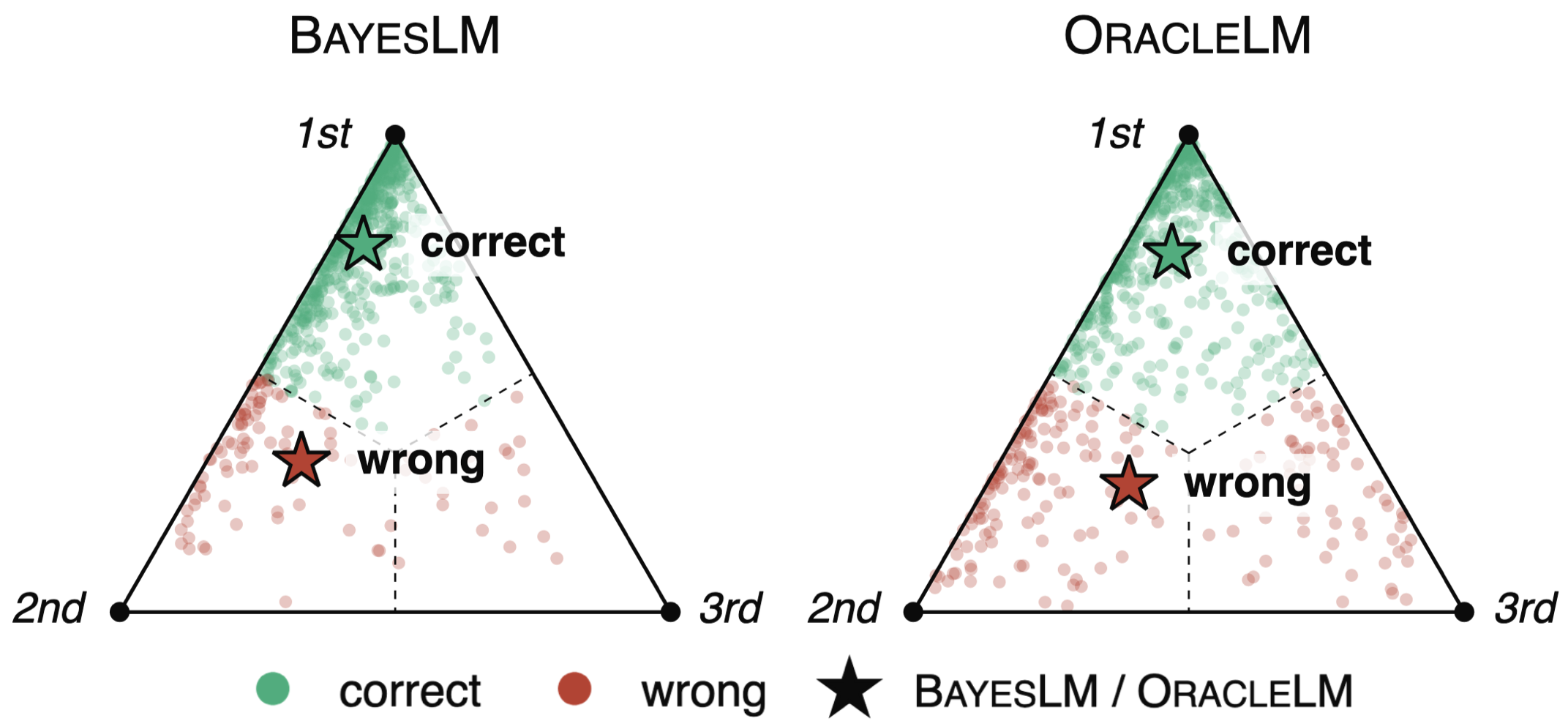}
      \caption{LMs' policies (dots) and their average (star) on the \textit{original} set, in the simplex based on \bayesianassistant{} rankings (top corner: most probable choice), colored by Bayes-correct (green) or not (red).}
      \label{fig:behavioral-simplex}
  \end{minipage}
\end{figure*}

\paragraph{Bayes accuracy is consistent across test sets.}
We compare the evaluated LMs with the \bayesianassistant{} behaviorally across all four test sets.
While we expect no difference between \textit{original} and \textit{shuffled} for \bayeslm, we do expect that noisy evidence should decrease its Bayes accuracy, but only under \textit{high-noise}. 
In line with \citet{qiu2026bayesian}, we also expect that \oraclelm{} is less Bayes-accurate but have no expectations about its Bayes-consistency. 
We test for significant effects by regressing Bayes correctness on the main effects of the LM, the test set, and their interaction.\footnote{For all regression tests in this paper we apply $p$-value correction using Bonferroni-Holm \citep{Holm1979ASS} with a significance level $\alpha = 0.05$, reporting all details in \autoref{app:stats}.} 

Results on the \textit{original} set in \autoref{fig:behavioral-accuracy} show that \bayeslm{} achieves a significantly ($p<10^{-4}$) higher Bayes accuracy of $0.84$ than \oraclelm{} ($0.69$), replicating the results in \citet{qiu2026bayesian}.
For \bayeslm{} and \oraclelm{}, we find that the accuracy did not significantly differ when comparing results on \textit{original} with \textit{shuffled} or \textit{low-noise} sets ($p>0.05$).
Also, the accuracy on the \textit{high-noise} set is significantly lower for them  ($p=0.022$).
\baselm, in contrast, achieves near-chance accuracy across all sets.
These results show that both \bayeslm{} and \oraclelm{} are Bayes consistent across test sets at the level of the best action (L1), as R1 requires, but Bayesian fine-tuning allows training a model more capable of reasoning under uncertainty than oracle fine-tuning. 

\paragraph{Policy divergence is Bayes-coherent.}
After evaluating the best action (L1), we turn to the choice policy (L2), derived from the LMs' token probabilities, and ask whether it aligns with the Bayesian policy.
We calculate the policy divergence for each model, and expect the \bayeslm{} divergence to stay lower than that of \oraclelm{}. 
The LMs' policies are more decisive, i.e., place more probability mass on the chosen option than the assistant, so we report results for the softmax parameter $\beta=4$, and provide parameter fitting details and discussion in \autoref{app:acting-bayesian}. 
The divergence on instances where the models' predictions were Bayes-correct is $0.10$ for \bayeslm{}, and $0.26$ for \oraclelm{}. 
On incorrect predictions, the $\beta=4$ scaling reduced the divergence only for \bayeslm{} (from $0.42$ to $0.38$) and increased it for \oraclelm{} ($0.66$ to $0.90$), suggesting that the LMs differ in the Bayes-consistency of the policies that lead to behaviorally incorrect predictions.

Since the Bayesian policy is a softmax over expected utilities (\autoref{sec:setup}), we expect an LM's wrong recommendation to be the \bayesianassistant{}'s second choice rather than its last.
\autoref{fig:behavioral-simplex} confirms this for \bayeslm{}.
When it fails, its choice policy accounts for the second most probable choice of the assistant (left corner) with an average probability of $52\%$, and the least-probable choice (right corner) with $17\%$.
In contrast, \oraclelm{} fails more uniformly with an average probability of $48\%$ and $26\%$ on the 2nd and 3rd choices. 
We expect Bayes-consistent failures to arise particularly when the Bayesian policy prefers no clear single choice, i.e., has higher entropy. 
To test this expectation, we regress the LMs' Bayes correctness on the entropy of the \bayesianassistant{}'s policy, the factor LM and their interaction. 
Indeed, both fine-tuned models fail more often when the assistant's policy is more uniform ($\beta = -0.49$, $p < 10^{-4}$), and the interaction shows this dependence is stronger for the \bayeslm{} ($\beta = -0.62$, $p = 0.014$).
The \bayeslm{}'s policy entropy also correlates significantly ($p < 10^{-4}$) more strongly with the assistant's (Pearson $r = 0.61$) than the \oraclelm{}'s does ($r = 0.30$), showing that it better captures \bayesianassistant{}'s uncertainty.

In sum, \bayeslm{} selects the best-action (L1) significantly more Bayes-accurately than \oraclelm{}, but both models pass consistency criteria across different test sets. 
We see that the difference is driven by the choice policy (L2), where \bayeslm{} shows a lower policy divergence than \oraclelm{}, and fails in a more Bayesian way. 
Behavior alone, however, cannot tell whether LMs encode beliefs that are consistent with the assistant, which is the question we ask next.


\section{Representing Beliefs: Probing Analysis}
\label{sec:representing-beliefs}
As incorrect beliefs can potentially lead to the same policy as Bayes-consistent beliefs, a model can agree with the \bayesianassistant{} at L1 and L2 while diverging from it at L3.
We therefore turn to \textbf{R2} and ask whether theoretically relevant quantities---prior, likelihood, and posterior---, as well as the policy, are encoded in the model's hidden states in a Bayes-consistent way.

\begin{figure*}[t]
  \centering
  \includegraphics[width=\textwidth]{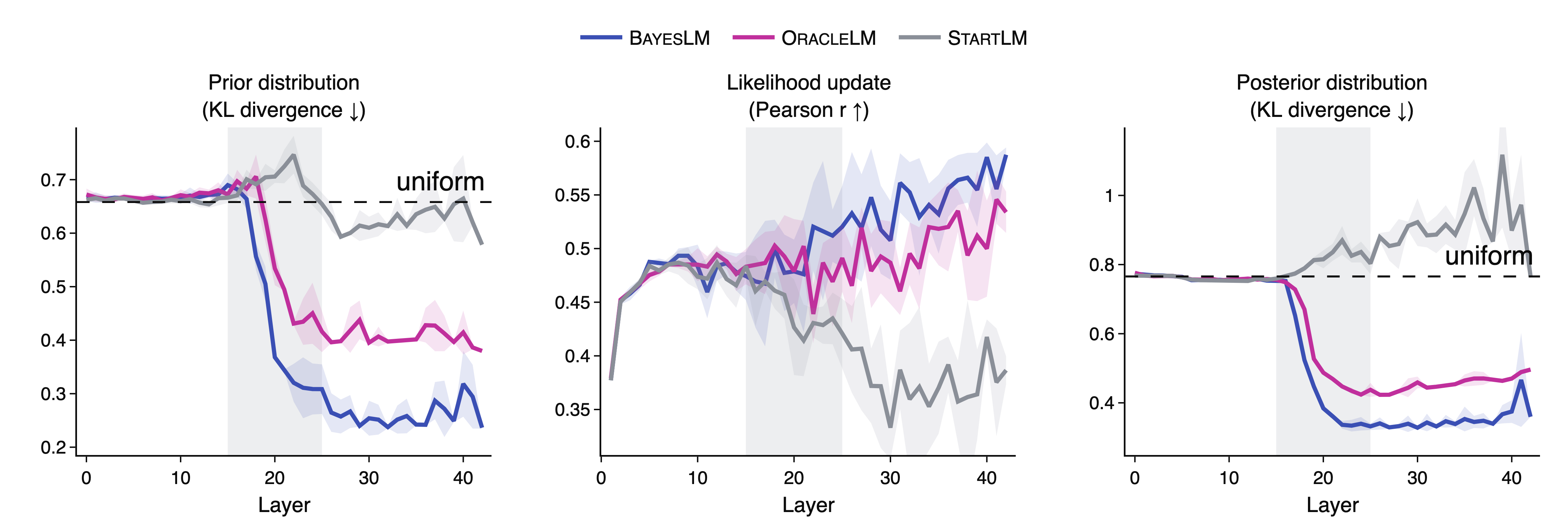}
  \caption{\textbf{The quantities of Bayes' rule emerge in the middle layers.}
  Layer-wise probing for the prior, the likelihood update, and the posterior, each scored against the \bayesianassistant{}.
  Dashed lines mark the reference KL between a uniform distribution and the assistant; the grey area marks the layers $15$--$25$ with the main effect; the bands show the deviation across the $20$ probes per layer.}
  \label{fig:probing-overview}
\end{figure*}

\paragraph{Decoded beliefs are consistent with the assistant.}
We test whether the prior, likelihood and posterior of the \bayesianassistant{} can be retrieved via linear probes from an LMs' hidden state $h_{\ell,t} \in \mathbb{R}^{d_{\text{model}}}$ at token position $t$ and layer $\ell \in \{0, \dots, 42\}$ at the final round~$5$ from the fine-tuning data of \citet{qiu2026bayesian}.
We read out at three positions: the \textit{feedback} token of round~$4$, the \textit{end of input}, and the \textit{digit} position that predicts the flight-number token (see \autoref{fig:prompt}).
Focusing on $\text{belief}_4$ (prior) and $\text{belief}_5$ (posterior) over each feature $j$ with five possible values, we fit distributional probes with a Kullback--Leibler (KL) loss and report the belief divergence as defined in \autoref{sec:setup}.
We approximate the likelihood term via the magnitude of the belief update, computing the labels for the probes as the penultimate belief update of the \bayesianassistant{} as $\text{belief}_4 - \text{belief}_3$, and the the LM's representation as the difference between the hidden layers at feedback tokens of the respective turns $h_{l,t_4} - h_{l,t_3}$.
We fit a regression probe 
and report its Pearson correlation with the true update.

If \bayeslm{} differs from \oraclelm{} not only superficially, we expect it to encode these quantities more strongly (i.e., lower divergences). 
\baselm{} provides a reference point for what a probe achieves with presumably little to no signal.
Indeed, all three quantities become decodable from layers~$15$--$25$ on in fine-tuned models (\autoref{fig:probing-overview}).
For \bayeslm{}, the divergence for $\text{belief}_4$ and $\text{belief}_5$ drops to $0.25$ and $0.35$ at the best layers, and the correlation with the likelihood update rises to $r=0.59$.
The \oraclelm{} shows the same pattern more weakly, with best values of $0.40$, $0.45$, and $0.53$, while the \baselm{} shows none of it.
As in \autoref{sec:acting-bayesian}, beliefs of both \bayeslm{} and \oraclelm{} are order-invariant and respond adequately to noisy input, suggesting that they mainly differ in how strongly they encode beliefs, while both being Bayes-consistent (see \autoref{app:probing-features}).

\begin{wrapfigure}[20]{r}{0.36\textwidth}
  \vspace{-\baselineskip}
  \centering
  \includegraphics[width=\linewidth]{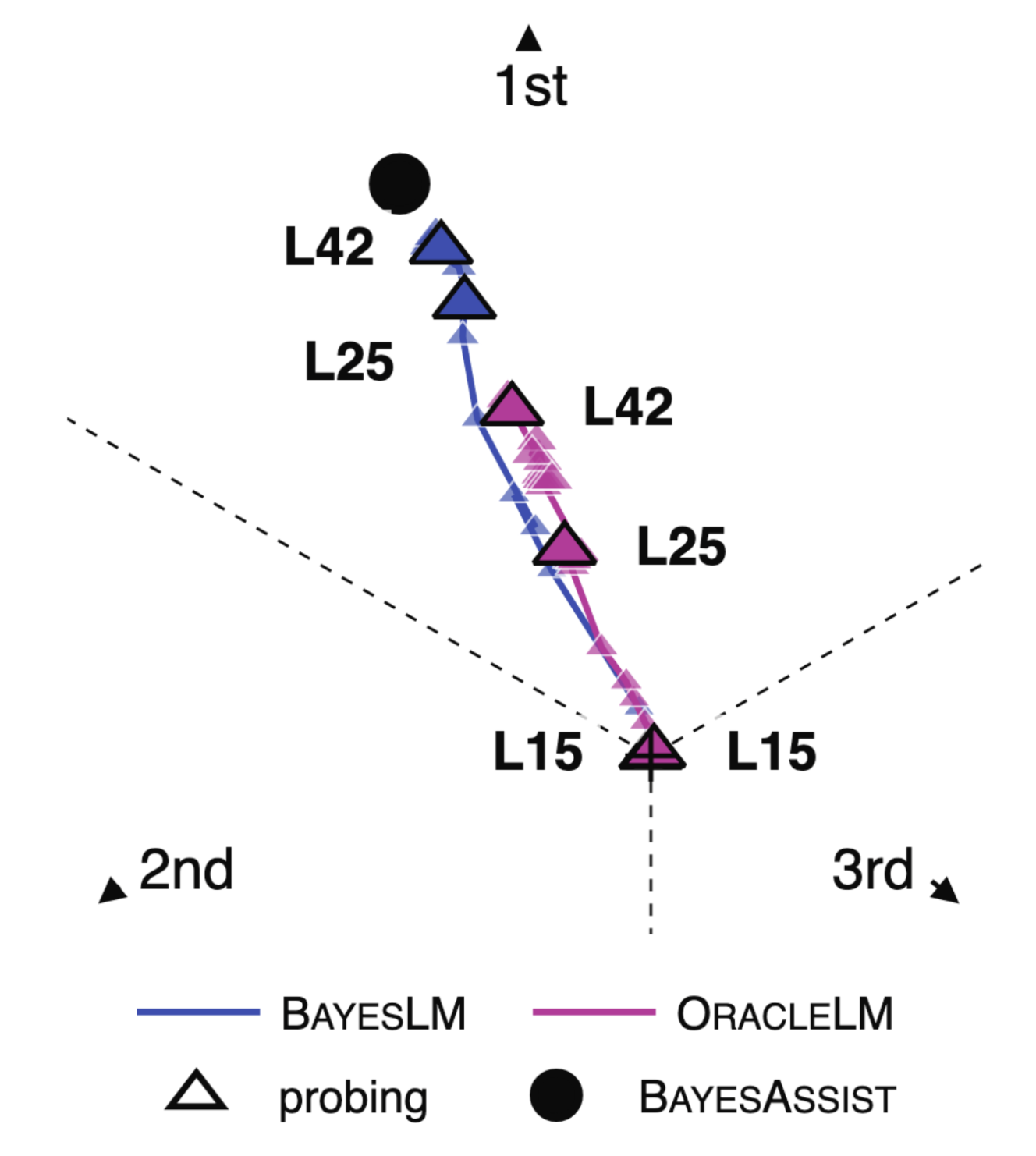}
  \caption{Path of the average decoded choice policy ($\triangle$) of \bayeslm{} and \oraclelm{} and the optimal one under the \bayesianassistant{} ($\bullet$ top corner).}
  \label{fig:choice-simplex}
\end{wrapfigure}
\paragraph{Analyzing the Bayes-consistency of decoded policies.}
Bayes-consistency at L2 asks for more than the beliefs being present, since the policy those beliefs imply must be encoded as well.
We therefore probe the choice policy over the three flight options with the same distributional setup, read out at the digit position (cf.~\autoref{fig:prompt}). 
\autoref{fig:choice-simplex} shows the average of the predicted distributions, which start uniform (center of simplex) around layer~$15$ and leave it during the middle layers, in the same window in which the beliefs arise.
Over all conversations, the \bayeslm{} comes close to the \bayesianassistant{}'s choice policy ($\beta=1$) by layer~$42$, while the \oraclelm{} stops before.
Accordingly, the Bayes accuracy of the decoded policy (derived via argmax) rises from chance to $0.87$ for the \bayeslm{} and $0.71$ for the \oraclelm{}, and to only $0.37$ for the \baselm{}.


To derive the choice policy consistently with the softmax optimization of EU in Bayesian decision making (\autoref{sec:setup}), LMs need to encode at least the latent $\text{belief}_4$ about the user’s preference and the resulting choice policy over flight options.
Consequently, we may expect the divergence of LMs’ actual (token-based) policy to reflect the probing success of the LMs’ beliefs and decoded policy.
We test this by regressing the (actual) policy divergence on the probed policy (or belief) divergence, the model, and their interaction.
As expected, the (actual) policy divergence increases with the decoded policy divergence ($\beta = 0.19$, $p < 10^{-4}$), and the \oraclelm{}’s policy divergence is significantly worse than that of the \bayeslm{} ($\beta = -0.11, p < 10^{-4}$), without significant interaction ($\beta = 0.01, p= 0.59$).
In contrast, regressing the policy divergence against the decoded belief divergence has no significant effect of belief ($p > 0.5$).
The results are visualized in the Appendix (\autoref{fig:decoded-policy-simplex}).

In sum, these results provide evidence that the representations of \bayeslm{} are more Bayes-consistent (\textbf{R2}) than those of the \oraclelm{}, and for both, a better-decoded policy tracks the behavioral-policy divergence, and thereby subsequent Bayes accuracy, whereas a better-decoded belief does not.
However, this is not evidence that the belief does not matter.
A high-entropy policy (exactly when the models fail) could potentially come from an incorrect belief representation resulting in downstream errors, or be the result of correct computations with an option set where all options have close to equal expected utility under the current belief.
We therefore cannot tell from the probes whether the belief is causally involved in the model's predictions at all, and if how it is used differs between models.
We address these questions next by intervening on beliefs directly.
\section{Acting on Bayesian Beliefs: Intervention Analysis}
\label{sec:acting-on-beliefs}
A model can carry a Bayesian belief in its hidden states and still reach its answer by another route \citep{ravichander-etal-2021-probing}.
So if an LM is Bayesian in its decision-making, its recommendation must use that belief.
We test this by injecting a specific prior into the model \citep{vig2020investigating, todd2023function, 3692070.3692690} and checking whether the behavioral recommendation changes to the one the new belief implies (\textbf{R3}).

\paragraph{The recommendation follows the belief.}
To intervene on the $\text{belief}_4$ representation, we build a prototypical hidden state $v$ for each of the four features and each value direction (positive/negative) at a given layer, averaging its representations for $500$ representative samples (i.e., samples with biggest belief values for the respective feature under the \bayesianassistant{}) from the training data. 
We patch this prototype $v$ into the residual stream $h$ at the \textit{feedback} token and the next $32$ tokens, with $\alpha=0.5$ across $8$ subsequent layers as $h' = (1 - \alpha)* h + \alpha*v$, resulting in a patched $\text{belief}_4'$.
We use the same $v$ for all $32$ tokens, but a separate one per layer.
The intervention spans tokens and layers because we found that an edit at a single position and layer caused no behavioral effect, suggesting that the information is distributed  
(see \autoref{app:belief-intervention} for additional results).
We compare the edit against an edit with a random vector of the same norm.
If an LM's recommendation is computed from an encoded belief in a Bayes-consistent way, it should change if the recommendation changes under the \bayesianassistant{}'s policy resulting from $\text{belief}_4'$
, and stay the same otherwise.
We therefore report accuracy averaged equally across recommendation-change and no-change samples (balanced Bayes accuracy), then averaged over the eight feature-direction settings.

At layers~$18$--$25$ the \bayeslm{} reaches a balanced accuracy of $0.77$ (visualized in \autoref{fig:patching-results} in the Appendix), and the intervention works across test sets (see \autoref{app:belief-intervention}).
The random vector barely changes the answer at all---so the injected belief, not the disturbance through patching itself, drives the model's recommendations.
The effect is confined to the layers where the belief becomes decodable, peaking around layers $15$--$22$ and falling back to chance in the upper layers.
Since the $\text{belief}_4$ remains decodable in those upper layers without being steerable there, this underscores that decodability alone does not imply a causal role.
\oraclelm{}'s accuracy of $0.52$, on the other hand, remains close to chance, suggesting that it does not \textit{use} its encoded beliefs.

\paragraph{Forming Bayesian beliefs relies on distributed information.} 
Based on the observation that a large-scale intervention is necessary to produce recommendation changes, we speculate that it might be because 
of the computational route for Bayes' rule in LMs.
Concretely, Bayesian inference can equivalently be realized with \textit{sequential} or \textit{one-swoop} accommodation of evidence for calculating the posterior, where the former would require maintaining intermediate posteriors, and the latter gathering information across context.
We offer a novel exploration whether the LM exhibits higher consistency with one of these routes.
We do so by applying masks that allow attention flow consistent with only one of the routes (details are in \autoref{app:masking}).
Results reveal a substantial drop from original Bayes accuracy of $0.84$ to $0.68$ under the mask allowing one-swoop, and to $0.56$ under the mask allowing sequential flow. 
This suggests that the \bayeslm{} indeed tends to follow a distributed, more one-swoop-like implementation, although both masks lead to accuracy drops.

In sum, \bayeslm{} partially (not reaching ceiling) uses its encoded beliefs for its recommendation (\textbf{R3}), while \oraclelm{}'s predictions are less sensitive to interventions on beliefs, so the models differ in the information their recommendations use.
However, the interventions do not show which \textit{computations} form the recommendation in the LM, which we address next.

\section{Forming a Bayesian Policy from Beliefs}
\label{sec:bayesian-implementation}
Without investigating the computations which use the represented beliefs, it remains unclear to what extent the quality of the belief representation (L3) affects the recommendation behavior in the different LMs: divergence between LM policies and \bayesianassistant{} could be, e.g., because of miscalibrated belief representations, or because correct beliefs are used inconsistently with Bayes-optimal decision making. 
So next we focus on \textbf{R4} and provide exploratory analyses of how the LMs' behavioral policy might be formed with the represented beliefs  (L3$\rightarrow$L2). 

\paragraph{Cross-model patching.}
In \autoref{sec:representing-beliefs} we showed that \bayeslm{}'s representations are more Bayes-consistent than other LMs', and that only \bayeslm{}'s beliefs are causally involved in the recommendation task (\autoref{sec:acting-on-beliefs}).
If the latter is simply due to more task-relevant, accurate content of \bayeslm{}'s beliefs at L3, we expect that patching \bayeslm{}'s beliefs into another model will increase its balanced accuracy on the tests from~\autoref{sec:acting-on-beliefs} up to the level of \bayeslm{}, 
while patching another model's beliefs into \bayeslm{} will lower it.
If additionally there is a difference in how the Bayes-fine-tuned LM \textit{forms} the recommendation from the belief representation (L3$\rightarrow$L2), we expect other LMs patched with \bayeslm{}'s beliefs to still underperform relative to \bayeslm{}.
To test this, we inject the belief prototype vector $v$ derived from one model into another, 
and compare the resulting accuracy against the results from injecting the model's own prototype belief.
We use the best hyperparameter setting from \autoref{sec:acting-on-beliefs}. 
\begin{figure*}[t]
  \centering
  \begin{minipage}[b]{0.45\textwidth}
    \centering
    \includegraphics[width=\linewidth]{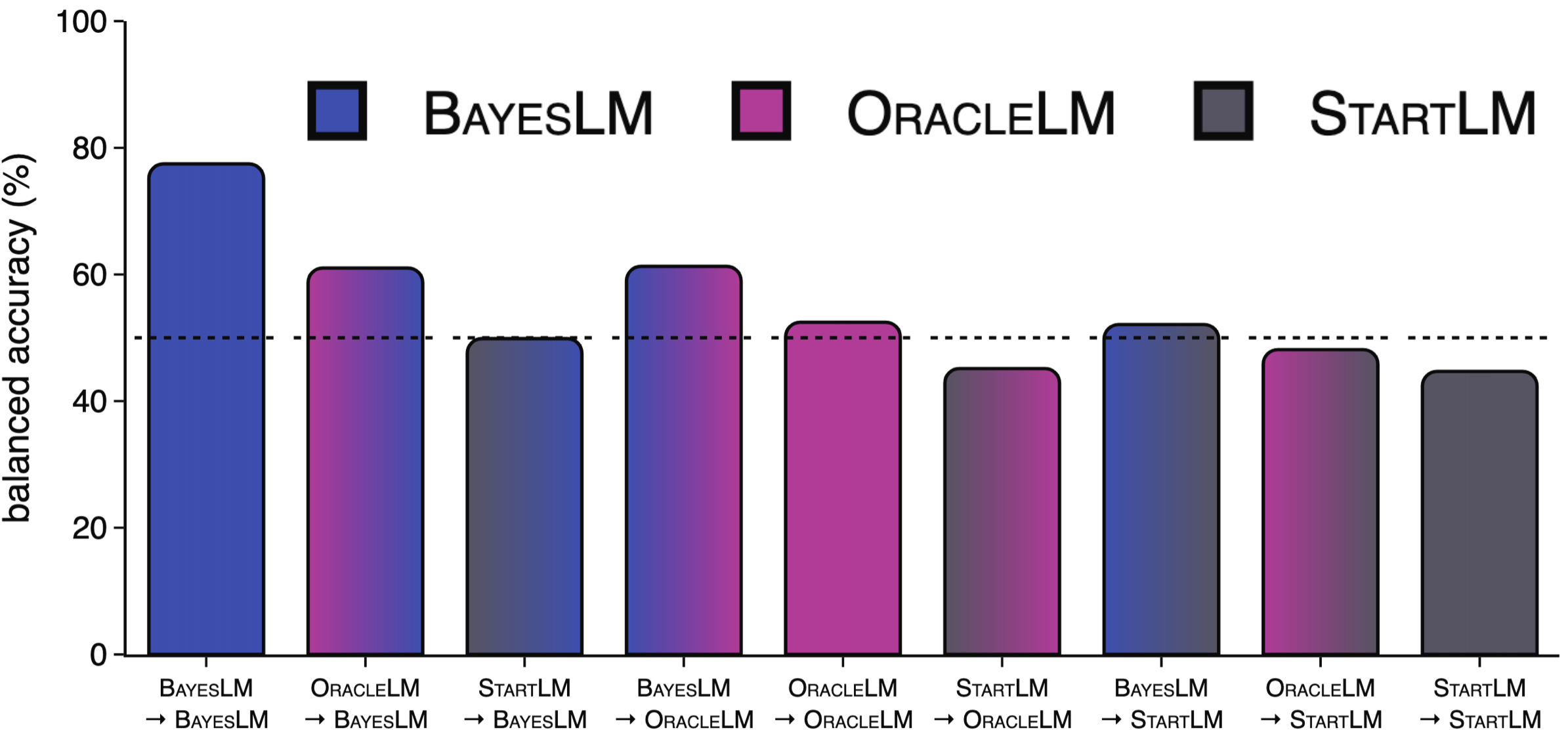}
    \captionof{figure}{
    Balanced accuracy when different models' prototype vector is patched into another models, relative to patching each model's own prototype vector.}
    \label{fig:cross-model-patching}
  \end{minipage}
  \hfill
  \begin{minipage}[b]{0.5\textwidth}
    \centering
    \begin{tabular}{lccc}
    \toprule
      & neuro- & $\Delta$ over & KL(NeSy$\mid \mid$ \\
      & symbolic & LM & LM)\\
    \midrule
    \bayeslm{}  & 0.86 & $0.02$ & 0.106\\
    \oraclelm{} & 0.78 & $0.09$ & 0.354\\
    \bottomrule
    \end{tabular}
    \captionof{table}{Bayes accuracy of neuro-symbolic against each model's own predictions, and divergence between neuro-symbolic and LMs' actual policies, representing what each model loses by its own belief read-out.}
    \label{tab:nesy}
  \end{minipage}
\end{figure*}

\autoref{fig:cross-model-patching} shows that \bayeslm{} beliefs raise the \oraclelm{}'s balanced accuracy from $0.52$ to $0.61$, while \oraclelm{} beliefs lower the \bayeslm{}'s from $0.77$ to $0.61$.
This confirms that \bayeslm{} supplies higher-quality, task-relevant beliefs (L3), but their content alone does not explain the gap between the models, because the patched \oraclelm{} still underperforms.
We address this next.

\paragraph{Deriving a neuro-symbolic behavioral policy.}
Previous results alone could not tell how closely the LMs' use of represented beliefs for \textit{behavioral} predictions matches Bayesian decision making, because it is not feasible to map the LMs' forward pass transforming belief representations into the behavioral policy onto the normative symbolic calculation of the softmax-rational choice policy.
To approach this calculation, we take the LMs' probed $\text{belief}_4$ (\autoref{sec:representing-beliefs}), calculate the posterior mean for each feature, and plug them into the symbolic EU calculation of the \bayesianassistant{} (\autoref{sec:setup}), resulting in a \textit{neuro-symbolic} model.
Results in Table~\ref{tab:nesy} show that the neuro-symbolic accuracy is nearly identical to \bayeslm{}'s own (+0.02), showing that \bayeslm{}'s forward pass already maps its beliefs close to the normative upper bound permitted by the quality of the encoded beliefs.
The neuro-symbolic model based on the \oraclelm{}, on the other hand, improves over the \oraclelm{} by $0.09$, suggesting that \oraclelm{} partially encodes Bayes-relevant beliefs without fully reading them out for recommendation.
Both neuro-symbolic policies diverge little from the Bayesian policy ($0.009$ for \bayeslm{}, $0.017$ for \oraclelm{}) but more from the LMs' own behavioral policies ($0.106$ and $0.354$), suggesting that the behavioral gaps in \autoref{sec:representing-beliefs} stem from noise in policy components other than beliefs (e.g., option-set representations), and more so for \oraclelm{}.\footnote{We apply beta scaling to the neuro-symbolic policy as in \autoref{sec:acting-bayesian} to calculate divergence to the LM behavioral policy.}
Notably, since the \bayeslm{} neuro-symbolic performance is also not at ceiling yet, we speculate that remaining behavioral differences between the \bayeslm{} and \bayesianassistant{} might be partially due to noisy belief decoding caused by distributed information in the LM.

In sum, we conclude that Bayesian fine-tuning installs not only a more consistent probabilistic internal representation, but also a read-out that at least in part consumes what the model encodes for the downstream task in a way that standard oracle fine-tuning cannot.
\section{Related Work}
\label{sec:related-work}

Bayesian inference in connection with LMs is studied in cognitive science \citep{griffiths2024bayes, zhu2026computation} and machine learning \citep{mackay1995probable, shi2009neural, mandt2017stochastic, papamarkou2026position}.
This includes analyzing probabilistic reasoning in general \citep{nafar2024probabilistic, samanta2026bayesbench}, on tasks with explicitly presented probabilities, showing that LMs may deviate from normative behavior \citep[e.g.,][]{paruchuri2024odds, zhu2024incoherent, pournemat2025reasoning}, and improving reasoning by external scaffolding \citep{piriyakulkij2024doing, feng2025bird}.
Additional work examines whether LMs can verbalize or interpret expressions of uncertainty \citep{belem-etal-2024-perceptions, tang2026evaluation}, their inferences about a choice policy \citep{liu2024large}, and divergence between human and model reasoning \citep{kamath2026humans}.
Relevant to our dataset, where the model infers preferences from sequential demonstrations \citep[cf.][]{lin2022inferring}, in-context learning is often viewed as approximate Bayesian inference (\citealp{xie2021explanation, zhang2023and, panwar2024context, gupta2025enough}; but see \citealp{falck2024context}).
In contrast to studies that focus on probabilistic reasoning through model behavior alone, or on model internals alone, we build on \citet{qiu2024can, qiu2026bayesian} and assess LMs on a task that requires both Bayesian inference and subsequent optimal decision-making using the inferred latent variable, across the behavioral and the internal perspective.
Studying model internals typically involves probing information in LMs' representations \citep[e.g.,][]{tenney-etal-2019-bert, waldis-etal-2024-holmes} or steering their behavior to uncover causal mechanisms \citep[e.g.,][]{turner2025steering, todd2023function}.
Related to this work, others have shown that representations encode the dynamics of beliefs on other tasks \citep{bortoletto2024brittle, bigelow2025belief, guiomar2026reasoning}, while the resulting behavior remains brittle, consistent with findings that probed information is not necessarily used by the model \citep{gekhman2025inside, orgad2025llms, waldis2026instructions}.

\section{Discussion}
\label{sec:discussion}

Results across R1--R4 (\autoref{tab:results-summary}) show that Bayesian fine-tuning makes an LM as Bayesian as its beliefs allow, but not a fully genuine Bayesian decision maker.
At \textbf{R1} and \textbf{R2}, \bayeslm{} acts Bayesian and represents the quantities of Bayes' rule.
At \textbf{R3}, it uses its decoded beliefs causally, but not as fully as Bayesian decision making would require.
At \textbf{R4}, its transformation of beliefs into a choice policy across options by expected utility is as good as those beliefs allow.
So the gap to the \bayesianassistant{} lies in the beliefs, not in their read-out.
Across all four requirements, \bayeslm{} is more consistent with the \bayesianassistant{} than \oraclelm{}, a pattern that holds across backbones (\autoref{app:backbones}).
These results have three implications: for how we fine-tune LMs, how we evaluate them, and what we can expect from them. 
We discuss limitations of our study in \autoref{app:limitations}.

\begin{wraptable}{r}{0.40\textwidth}
\centering
\caption{Overview of results. Fine-tuning with Bayesian signal, but not oracle one, leads to stronger compliance with the requirements.}
\label{tab:results-summary}
\begin{tabular}{lcccc}
\toprule
  & R1 & R2 & R3 & R4 \\
\midrule
\baselm{}   & $\times$ & $\times$ & $\times$ & $\times$ \\
\oraclelm{} & \checkmark & \checkmark & $\times$ & $\times$ \\
\bayeslm{}  & \checkmark & \checkmark & (\checkmark) & (\checkmark) \\
\bottomrule
\end{tabular}
\end{wraptable}
\paragraph{The advantage of Bayesian supervision.}
As the explicit assistant can recover Bayes accuracy from the decoded beliefs of \bayeslm{} or \oraclelm{}, fine-tuning makes LMs act on most of what they encode.
But the beliefs of \bayeslm{} are superior, as patching its beliefs into \oraclelm{} clearly improves its Bayes accuracy.
These insights align with superficial alignment theory and the fact that fine-tuning rather impacts how LMs verbalize already encoded information than changing encoding itself \citep{zhou2023lima, waldis2026instructions}.
This has a direct implication for fine-tuning practice.
A training signal that reflects a Bayes-consistent uncertainty yields better reasoning under uncertainty than the standard protocol of tuning on an oracle-correct signal.

\paragraph{Behavior is not enough to understand reasoning under uncertainty.}
A wrong recommendation can come from a wrong belief or from how a right belief is transformed into a recommendation across options.
Behavioral evaluation cannot distinguish between these, so it cannot tell whether a training change improved what a model encodes or how it uses the information, a fundamental gap between encoding and expression, also reflected in other work \citep{gekhman2025inside}.
This requires an analytical lens on decision-making, and we provide a first one with four concrete requirements and their operationalizations for LMs.


\paragraph{A Bayesian fine-tuned LM is still a language model.}
While the \bayesianassistant{} performs exact inference on each instance, LMs learn one forward pass that should serve all possible user tasks, and so may only approximate the posterior. They are trained to predict tokens rather than maintain uncertainty which explains their failures on high-entropy policies.
In addition, LMs compute beliefs across layers and tokens rather than at a single position, 
reading them out only in part.
However, in principle, since the target is learnable, more or better Bayesian fine-tuning may close the remaining gap.
How beliefs are formed and turned into a choice in the forward pass, however, remains open.

\section{Conclusion}
\label{sec:conclusion}
We asked to what extent an LM that behaves like a Bayesian decision maker also is one internally.
To answer this, we formulate four increasingly demanding requirements for Bayesian decision making in LMs, spanning behavior, representations, and computation, and applied them to the models and flight recommendation task of \citet{qiu2026bayesian}.
We show that \textbf{Bayesian fine-tuning makes an LM as Bayesian as its beliefs allow, but not a genuine Bayesian decision maker}.
Our requirements allow not only to detect whether an LM is Bayesian but also to locate where it stops being so, e.g., under oracle SFT, and they provide first insights into how an LM computes with beliefs.
It uses them causally, distributed across layers and tokens, and turns them into a choice policy as well as they allow.
While we lay groundwork for 
studying Bayesian decision making in LMs, future work should extend it in two directions.
First, to other tasks with a normative Bayesian solution, to test whether the pattern we find is specific to preference inference. 
Second, to the implementation of belief updating and of the transformation of beliefs into a choice policy across options \citep[cf.][]{piotrowski2025constrained}, given that the \bayeslm{}'s beliefs are distributed rather than held at a single position.

\subsection*{AI use statement}
In this work, we used generative AI tools for  coding the experiments, implementing analyses and visualizations, and for feedback on the interpretation of the results.
We have \textit{not} used generative AI tools for developing the conceptual framework, data generation or designing the experiments,
and the use of AI tools for mathematical proofs or for translation are not applicable to this work.
Additionally, we used generative AI tools for drafting parts of the text. We have reviewed all AI-assisted work. 
We take responsibility for the final content of this work, including text or artifacts produced with the aid of generative AI.
\subsubsection*{Acknowledgments}
We thank Ethan Wilcox for helpful discussions. 
PT is funded by the Deutsche Forschungsgemeinschaft (DFG, German Research Foundation) under project ID 579368432. 
PT and AW gratefully acknowledge the support by the state of Baden-W\"urttemberg through bwHPC and the German Research Foundation (DFG) through grant INST 35/1597-1 FUGG, and the data storage service SDS@hd supported by the Ministry of Science, Research and the Arts Baden-W\"urttemberg (MWK) and the German Research Foundation (DFG) through grant INST 35/1803-1 FUGG and INST 35/1804-1 LAGG.
AW and MF are supported by the Volkswagen Foundation through a Momentum grant.
TL's work on this project was supported by the National Science Foundation (NSF) under Cooperative Agreement No. 2433429, ``NSF AI Research Institute on Interaction for AI Assistants (ARIA)''.
MF is a member of the Machine Learning Cluster of Excellence at University of T\"ubingen, EXC number 2064/2 – Project number 39072764.
\bibliography{references}
\bibliographystyle{iclr2027_conference}

\appendix
\appendix

\section{Limitations}
\label{app:limitations}
\paragraph{Task scope.}
We compare models fine-tuned on a single task.
While it is a realistic assistant recommendation task, it is highly structured and depends on certain assumptions, e.g., the independence of reward features.
The ground truth \bayesianassistant{} generates the training data through softmax-rational decision making based on Bayesian inference over a latent user preference, and the analyzed models were trained on sequential, essentially in-context learning data.
Whether our findings hold for less structured tasks, or for tasks without an analytically solvable Bayesian model, is open.

\paragraph{Backbones.}
The main reported results focus on Gemma-2-9B-Instruct \citep{team2024gemma}.
Additional results for Qwen-2.5-7B and Llama-3.1-8B in \autoref{app:backbones} show qualitatively similar patterns, but we did not examine how architectural factors influence the success of Bayesian teaching, such as the lower Bayesian consistency of Qwen. Future work should additionally explore the interaction of Bayesian teaching and scaling, applying it to larger models.

\paragraph{Linear decodability.}
We use linear probes to test whether the models represent the quantities of Bayes' rule, which provides only a lower bound on the information contained in the LMs.
Our control tasks \citep{hewitt-liang-2019-designing} show that the probes are selective, and the \baselm{} serves as a further control: the same probes recover far less from it, so what we decode from the fine-tuned models is installed by fine-tuning rather than an artifact of probing.
Still, a non-linear read-out could recover more, and the belief divergence we report may partly reflect the limits of linear decoding rather than of the representation.

\paragraph{Scope of the interventions.}
Our causal interventions replace the prior only, at a fixed window of layers and tokens chosen for the \bayeslm{}.
They show that the beliefs are used, but not how they are combined, and the neuro-symbolic and cross-model results test the transformation into a choice policy from outside the model rather than by intervening on it.
We therefore lack an in-depth understanding of how an LM computes expected utility when transforming its beliefs into a choice policy across options.
Our results provide a first step in this direction.

\section{Statistical analyses}
\label{app:stats}
Below, we report details of the regression analyses performed in the main results.
For all regression models, we use z-scored normalized values for continuous predictors, and apply the Bonferroni-Holm $p$-value correction \citep{Holm1979ASS}.
The logistic regression model for Bayes accuracy in \autoref{sec:acting-bayesian} is reported in \autoref{tab:regression-acting-bayesian-accuracy}, the logistic regression for analyzing policy divergence is in \autoref{tab:regression-acting-bayesian-policy}. 
The linear regression for analyzing Bayes-consistency of the decoded policy in \autoref{sec:representing-beliefs} is in \autoref{tab:regression-representing-beliefs-policy}. A model with both predictors (both prior and decoded policy divergences) produced the same results.

\begin{table}[t]\centering\small
\caption{Logistic regression analysis of Bayes accuracy in \autoref{sec:acting-bayesian}, regressing Bayes correctness against the LM, the test set and their interaction: \texttt{correct $\sim$ C(model) * C(set)}. The predictors are dummy coded. OR refers to odds ratio.}
\label{tab:regression-acting-bayesian-accuracy}
\begin{tabular}{lrcrr}
\toprule
term & $\beta$ & 95\% CI of the OR & OR & $p$ \\
\midrule
BayesLM & $0.8816$ & $[1.8942 , 3.0781 ]$ & 2.4147 & $<10^{-14}$ \\
StartLM    & $-1.4257$ & $[0.1881, 0.3070]$ & 0.2403 & $<10^{-14}$ \\
Shuffled & $-0.0299$ & $[0.8231, 1.1444]$ & 0.9705 & $1.00$ \\
Low-Noise & $-0.2102$ & $[0.6620, 0.9923]$ & 0.8105 & $0.209$ \\
High-Noise & $-0.4215$ & $[0.5239, 0.8216]$ & 0.6561 & $0.002$ \\
BayesLM:Shuffled & $-0.0062$ & $[0.7899, 1.2503]$ & 0.9938 & $1.00$ \\
StartLM:Shuffled & $0.1133$ & $[0.9140, 1.3724]$ & 1.1200 & $1.00$ \\
BayesLM:Low-Noise & $0.3784$ & $[1.0370, 2.0555]$ & 1.4600 & $0.1809$ \\
StartLM:Low-Noise & $0.3141$ & $[1.0570, 1.7734]$ & 1.3691  & $0.1213$ \\
BayesLM:High-Noise & $-0.0306$ & $[0.6877, 1.3679]$ & 0.9699   & $1.00$ \\
StartLM:High-Noise & $0.5527$ & $[1.2711, 2.3763]$ & 1.7380  & $0.0043$ \\
\bottomrule
\end{tabular}
\par\smallskip
\end{table}

\begin{table}[t]\centering\small
\caption{Logistic regression analysis of policy divergence in \autoref{sec:acting-bayesian}, regressing Bayes correctness against the LM, the entropy of the Bayesian policy and their interaction: \texttt{correct $\sim$ C(model) * H(Bayesian policy)}. The model is dummy coded. OR refers to odds ratio.}
\label{tab:regression-acting-bayesian-policy}
\begin{tabular}{lrcrr}
\toprule
term & $\beta$ & 95\% CI of the $\beta$ & OR per SD & $p$ \\
\midrule
BayesLM & $1.1634$ & $[0.8004, 1.5264 ]$ & 3.2007 & $<10^{-14}$ \\
StartLM    & $-1.4757 $ & $[-1.7296, -1.2219]$ & 0.2286 & $<10^{-14}$ \\
H(policy)    & $-0.4949 $ & $[-0.7101, -0.2798]$ & 0.6096 & $<10^{-14}$ \\
H(policy):BayesLM    & $-0.6238$ & $[-1.1192, -0.1284 ]$ & 0.5359 & $0.0136$ \\
H(policy):StartLM    & $0.4290$ & $[0.1339, 0.7241]$ & 1.5357 & $0.0088$ \\
\bottomrule
\end{tabular}
\par\smallskip
\end{table}

\begin{table}[t]\centering\small
\caption{Linear regression analysis of Bayes-consistency of decoded beliefs in \autoref{sec:representing-beliefs}, regressing the LM's token-based behavioral policy against the LM, one of the two predictors (probed policy divergence or prior belief divergence) and their interaction. The model is dummy coded.}
\label{tab:regression-representing-beliefs-policy}
\begin{tabular}{lrcrr}
\toprule
term & $\beta$ & 95\% CI of the $\beta$ & $p$ \\
\midrule
\multicolumn{5}{l}{\texttt{div(behavioral policy) $\sim$ C(model) * div(prior)} ($R^2=0.02$, AIC=473.7)} \\
Intercept & $0.5037$ & $[0.4773, 0.5300]$ & --- \\
BayesLM    & $-0.0829 $ & $[-0.1117, -0.0541]$ & $<10^{-14}$ \\
div(prior)    & $0.0025 $ & $[-0.0250, 0.0300]$ & $1.00$ \\
div(prior):BayesLM    & $0.0107$ & $[-0.0237, 0.0450]$ & $1.00$ \\
\midrule
\multicolumn{5}{l}{\texttt{div(behavioral policy) $\sim$ C(model) * div(probed policy)} ($R^2=0.46$, AIC=-276.4)} \\
Intercept & $0.5144$ & $[0.4923, 0.5364 ]$ & --- \\
BayesLM    & $-0.1064 $ & $[-0.1283, -0.0846 ]$ & $<10^{-14}$ \\
div(policy)    & $0.1927 $ & $[0.1645, 0.2210]$ & $<10^{-14}$ \\
div(policy):BayesLM    & $0.0083$ & $[-0.0217, 0.0383]$ & $0.5880$ \\
\bottomrule
\end{tabular}
\par\smallskip
\end{table}

\section{Behavioral Evaluations}
\label{app:acting-bayesian}
We provide further details of the behavioral evaluations (R1: acting Bayesian) described in \autoref{sec:acting-bayesian} below.

\paragraph{\bayesianassistant{}'s performance.} 
The normative solution to the recommendation task is implemented by the \bayesianassistant{}, which makes recommendations based on beliefs that can be inferred from the available data of the preceding turns, which may not be fully disambiguating. Therefore, given the task data, the \bayesianassistant{}'s average Oracle accuracy (i.e., match with user's true preference) on the fifth round is 0.811 (\textit{original} set), 0.811 (\textit{shuffled} set), 0.683 (\textit{low noise} set), 0.420 (\textit{high noise} set).

\begin{figure}[t]
    \centering
    \includegraphics[width=0.8\linewidth]{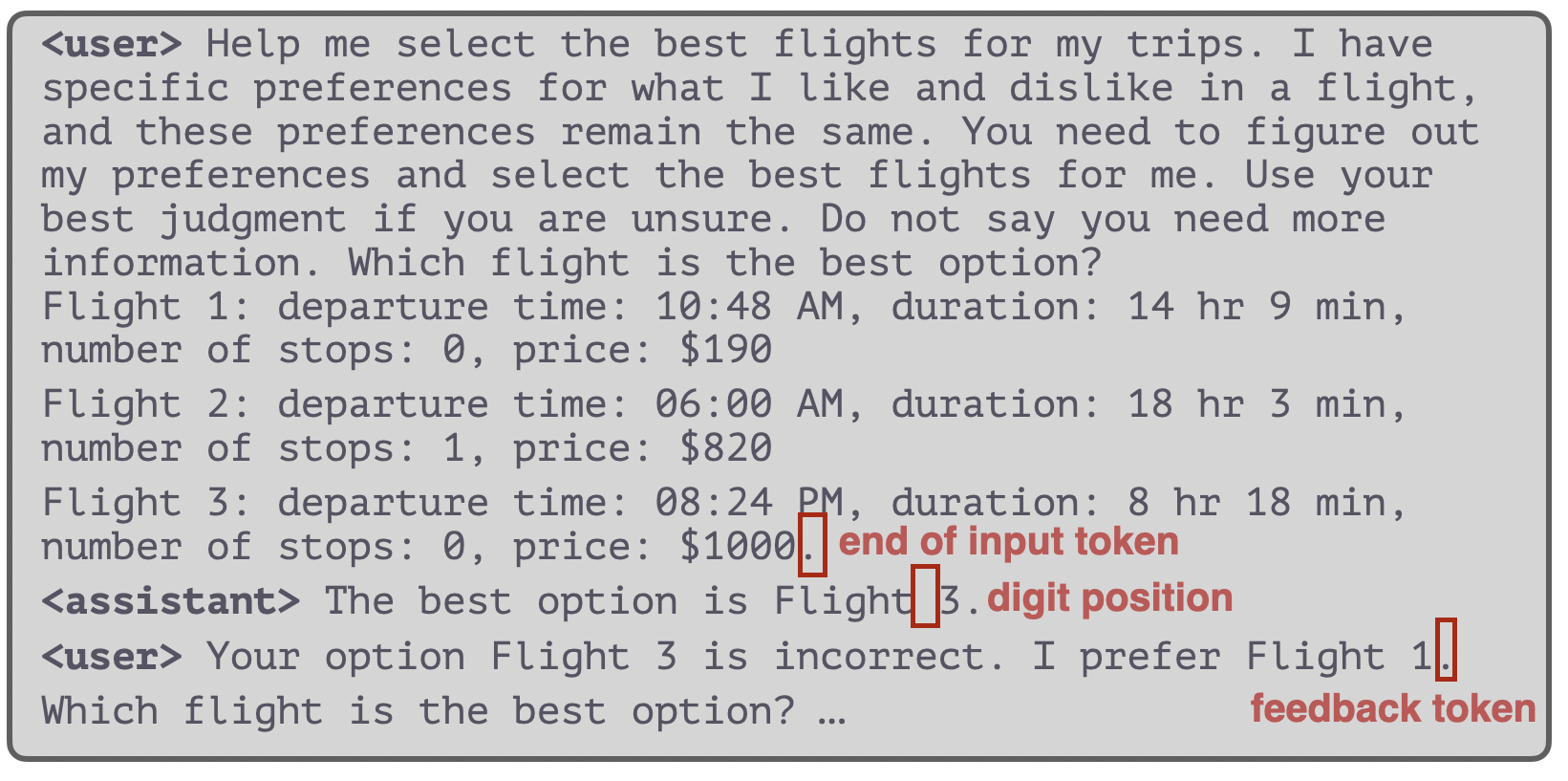}
    \caption{Prompt presenting the first round of the recommendation task, marking the probed token positions (reported probing was performed on round 5).}
    \label{fig:prompt}
\end{figure}
\begin{figure}[t]
    \centering
    \includegraphics[width=0.6\linewidth]{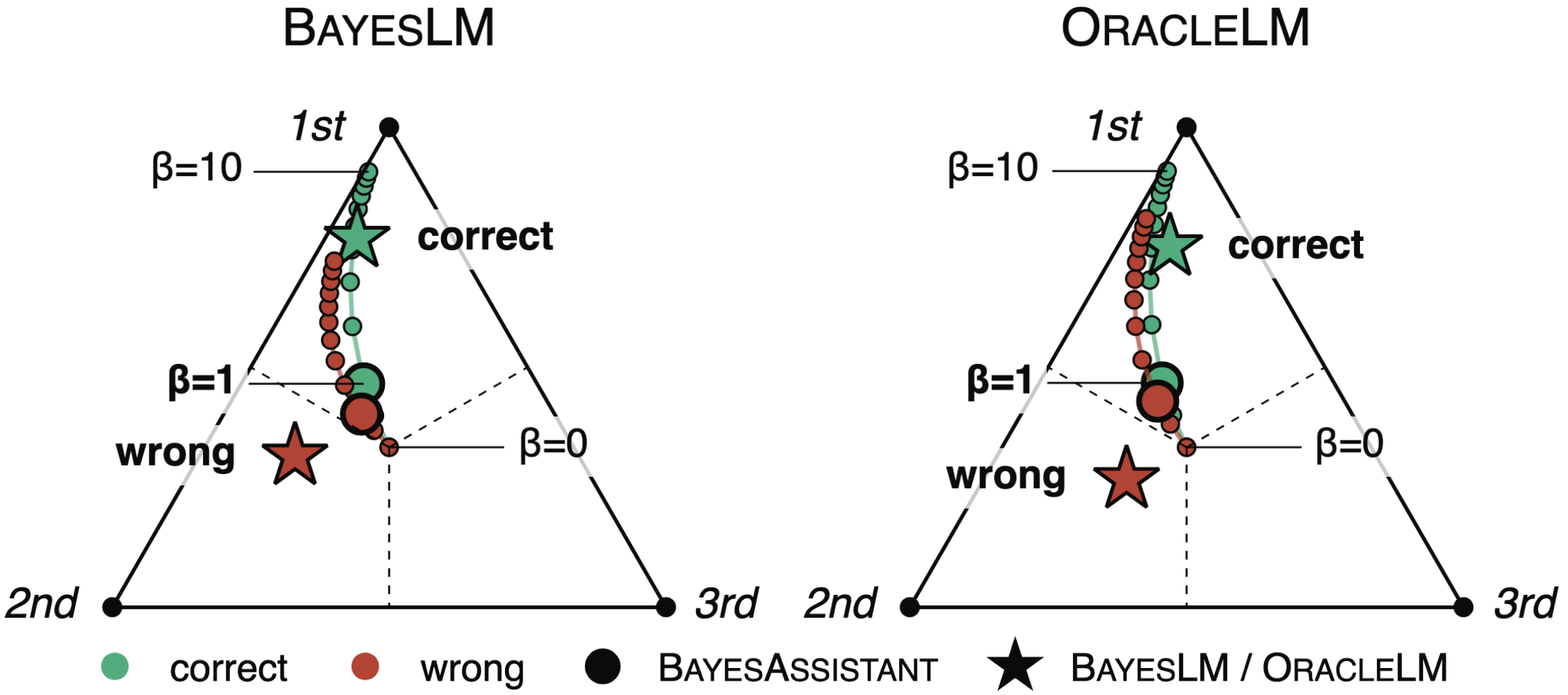}
    \caption{Visualized search over the softmax rationality parameter $\beta$ for scaling the \bayesianassistant{}'s policy: 
$\beta=0$ is a uniform distribution, $\beta=1$ corresponds to the original \bayesianassistant{}, $\beta=10$ is near-deterministic, and $\beta\approx4$ is the best match to the LMs' choice policies.}
  \label{fig:behavioral-simplex-param-search}
\end{figure}

\paragraph{Dataset details.}
We provide an example conversation round from the \textit{original} set in prompt format, as it was passed to the LMs, in \autoref{fig:prompt}.
Following \citet{qiu2026bayesian}, the dataset features departure time, duration, and price each had ten distinct equally spaced values, and were mapped from the verbal representations onto equally spaced numbers between 0 and 1.0 for calculations within the \bayesianassistant{}. The stops feature only had three values (verbalized: 0, 1, 2; mapped onto 0, 0.5, 1, respectively). This was used across datasets.

\paragraph{Policy divergence analyses.}
We evaluate LMs' behavioral choice policies, derived from their conditional token probabilities, by calculating the KL divergence relative to the Bayesian (\bayesianassistant{}'s) policy (i.e., policy divergence, as introduced in \autoref{sec:setup}).
The Bayesian policy can be scaled by the rationality parameter $\beta$ (see \autoref{sec:setup}).
Under $\beta=1$, both LMs place more probability mass on the most probable option (i.e., emitted token, $\bigstar$ in \autoref{fig:behavioral-simplex-param-search}) than the \bayesianassistant{} does ($\Circle$ in \autoref{fig:behavioral-simplex-param-search}): while, on average, the \bayeslm{}'s chosen option has the probability $0.74$ and, \oraclelm{}'s $0.72$, the \bayesianassistant{}'s chosen option has the average probability $0.46$.
This indicates that LMs' policies are generally more decisive, suggesting that they may be less sensitive to nuanced uncertainty and may be more prone to errors.
For main comparisons we select a $\beta$ minimizing the policy divergence across LMs by considering integer $\beta$ values between 0 and 10 (the search is visualized separately for each LM in \autoref{fig:behavioral-simplex-param-search}).
The policy divergence is lowest for both LMs under $\beta= 4$: on samples where the models are Bayes-correct, the KL is $0.10$ for \bayeslm{}, and $0.26$ for \oraclelm{}. 
On samples with incorrect predictions, $\beta=4$ leads to a drop only for \bayeslm{} (KL of $0.42$ to $0.38$) and to an increase for \oraclelm{} ($0.66$ to $0.90$).

\begin{figure}[t]
    \centering
    \includegraphics[width=\textwidth]{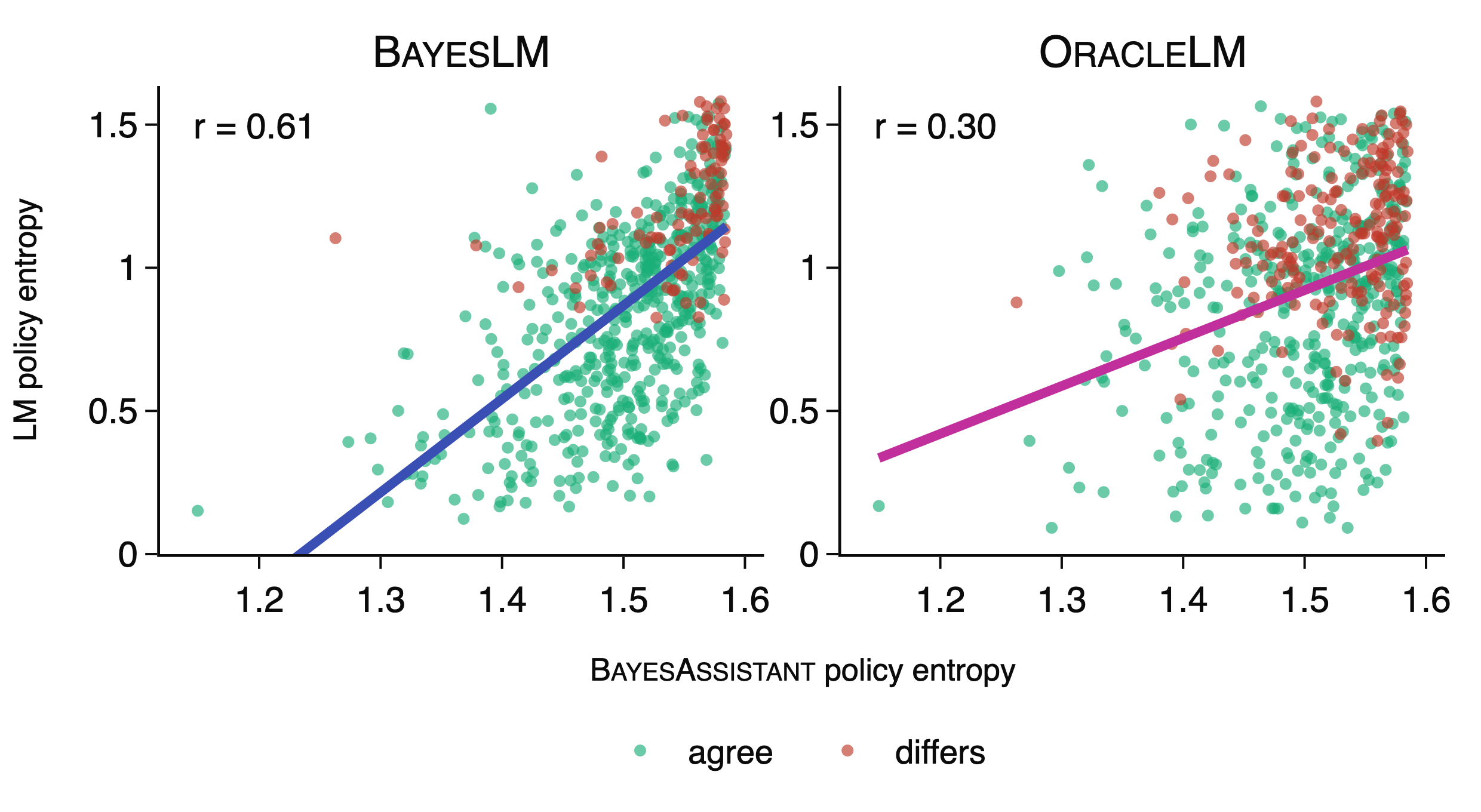}
    \caption{\textbf{The \bayeslm{}'s uncertainty tracks the assistant's.}
    Entropy of the LM's behavioral choice policy against the entropy of the Bayesian policy, one point per sample of the \textit{original} set, colored by whether the recommendation agrees with the \bayesianassistant{}.
    The \bayeslm{}'s entropy correlates with the assistant's ($r = 0.61$), the \oraclelm{}'s only weakly ($r = 0.30$).
    For both models, the differing instances are concentrated where the \bayesianassistant{} itself is most uncertain, but spread more for the \oraclelm{}.}
    \label{fig:policy-entropy-correlation}
\end{figure}

Additionally, we show the correlations of the entropies of the resulting Bayesian policy with the LM policies in \autoref{fig:policy-entropy-correlation}.

\begin{figure}[t]
    \centering
    \includegraphics[width=0.7\textwidth]{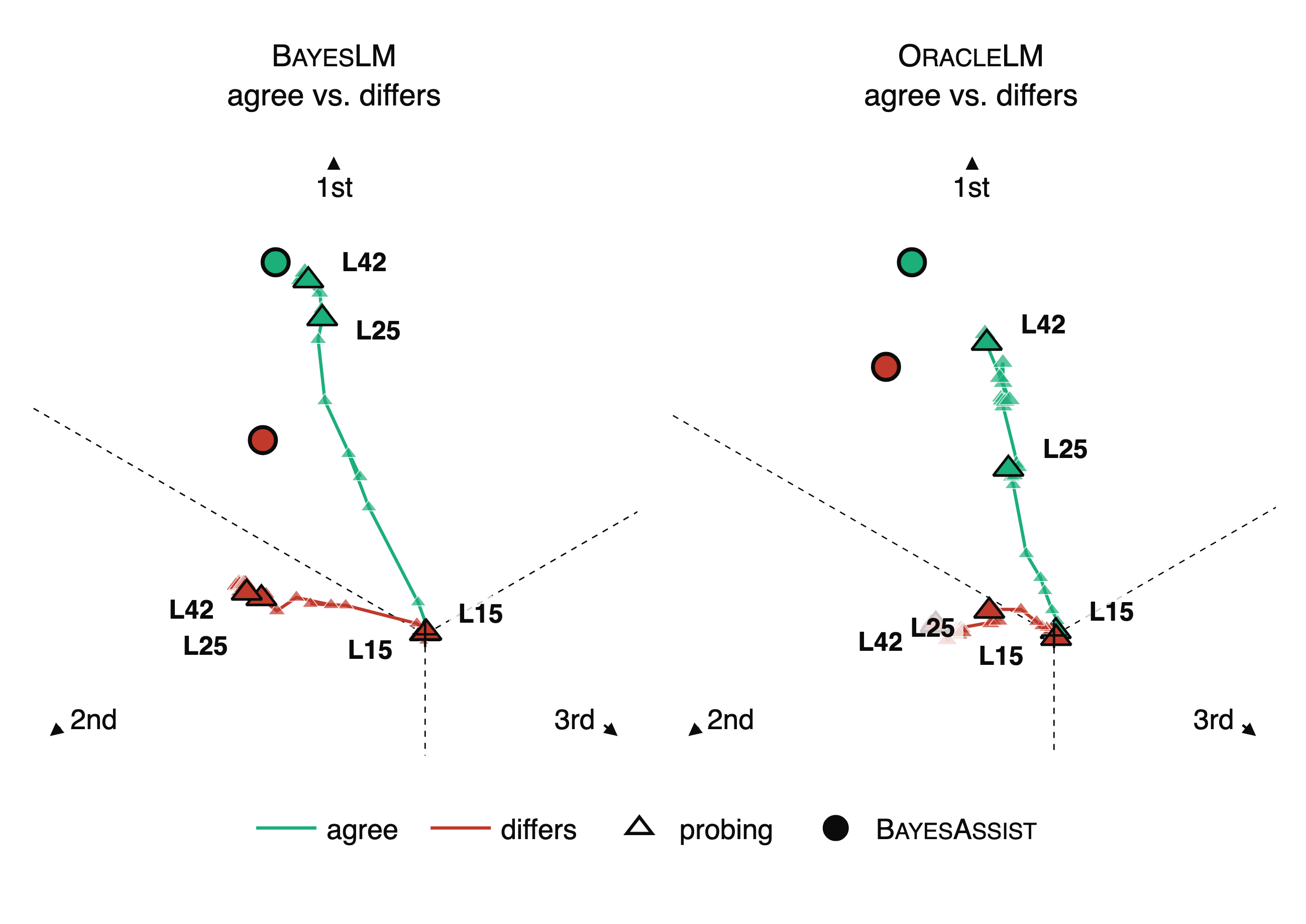}
    \caption{
    Mean decoded choice policy by layer on the simplex (corners ordered by Bayesian policy), split by whether the LM's recommendation agrees with the \bayesianassistant{}, i.e., is Bayes-correct, (green) or differs (red).
    Triangles mark the mean decoded choice policy for each layer, and circles mark the mean Bayesian policy of the \bayesianassistant{} for the same instances.}
    \label{fig:decoded-policy-simplex}
\end{figure}

\section{Probing Experiments}
\label{app:probing-features}
Below, more details of probing experiments from \autoref{sec:representing-beliefs} testing LMs' representations (R2: representing beliefs) are reported.
\paragraph{Additional results on Bayes-consistency of decoded policies.} In the main analyses, we identified that the probed policy divergence was predictive of the LMs' behavioral policy divergence. This is confirmed by visual inspection (see \autoref{fig:decoded-policy-simplex}), which shows that the decoded policy of \bayeslm{} comes close to the Bayesian policy, while the decoded policy of \oraclelm{} stays further away from it, in particular, on incorrect samples. 


\begin{figure*}[t]
  \centering
  \includegraphics[width=\textwidth]{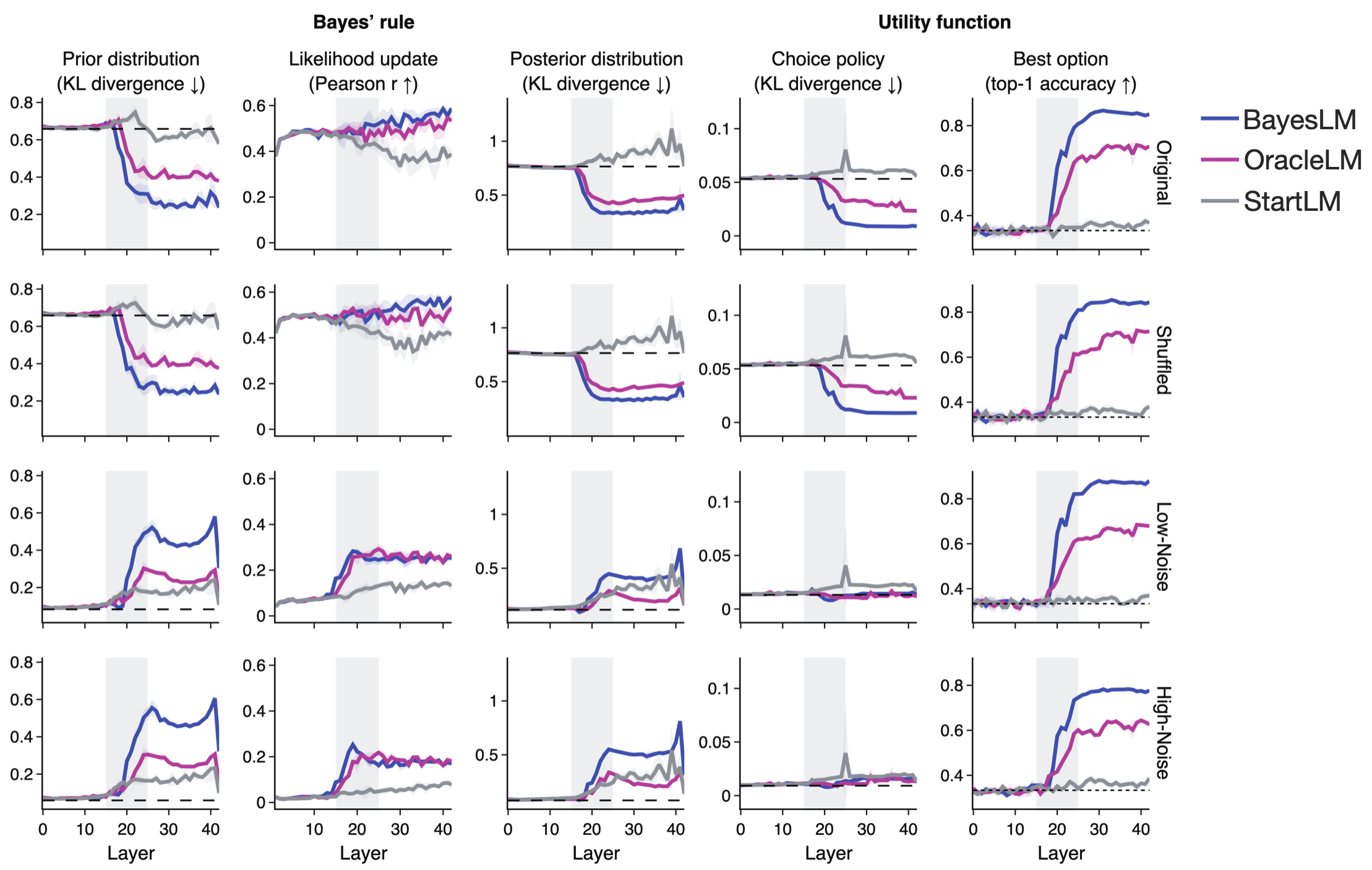}
  \caption{\textbf{Per-layer probing across the four evaluation sets (detail for
  \autoref{fig:probing-overview}).} Each row is an evaluation set (original,
  shuffled, low-noise, high-noise); columns first present probed quantities of Bayes' rule, and then the probed policy and resulting prediction (argmax of probed policy).
  On the original and shuffled sets, the belief and choice KL curves fall in the mid-stack and the \bayeslm{} outperforms \oraclelm{} and \baselm{} in the late layers. On the noisy sets the prior/posterior KL instead \emph{rises} in the upper layers: the probes are trained on the original set and learn to represent sharper beliefs in late layers, which diverges from the flatter beliefs induced by the noisy sets. Therefore, the KL grows even as choice top-1 stays high --- likely a calibration gap of the probe,
  not a representational failure of the model. }
  \label{fig:detailed-heldout}
\end{figure*}

\begin{figure*}[t]
  \centering
  \includegraphics[width=\textwidth]{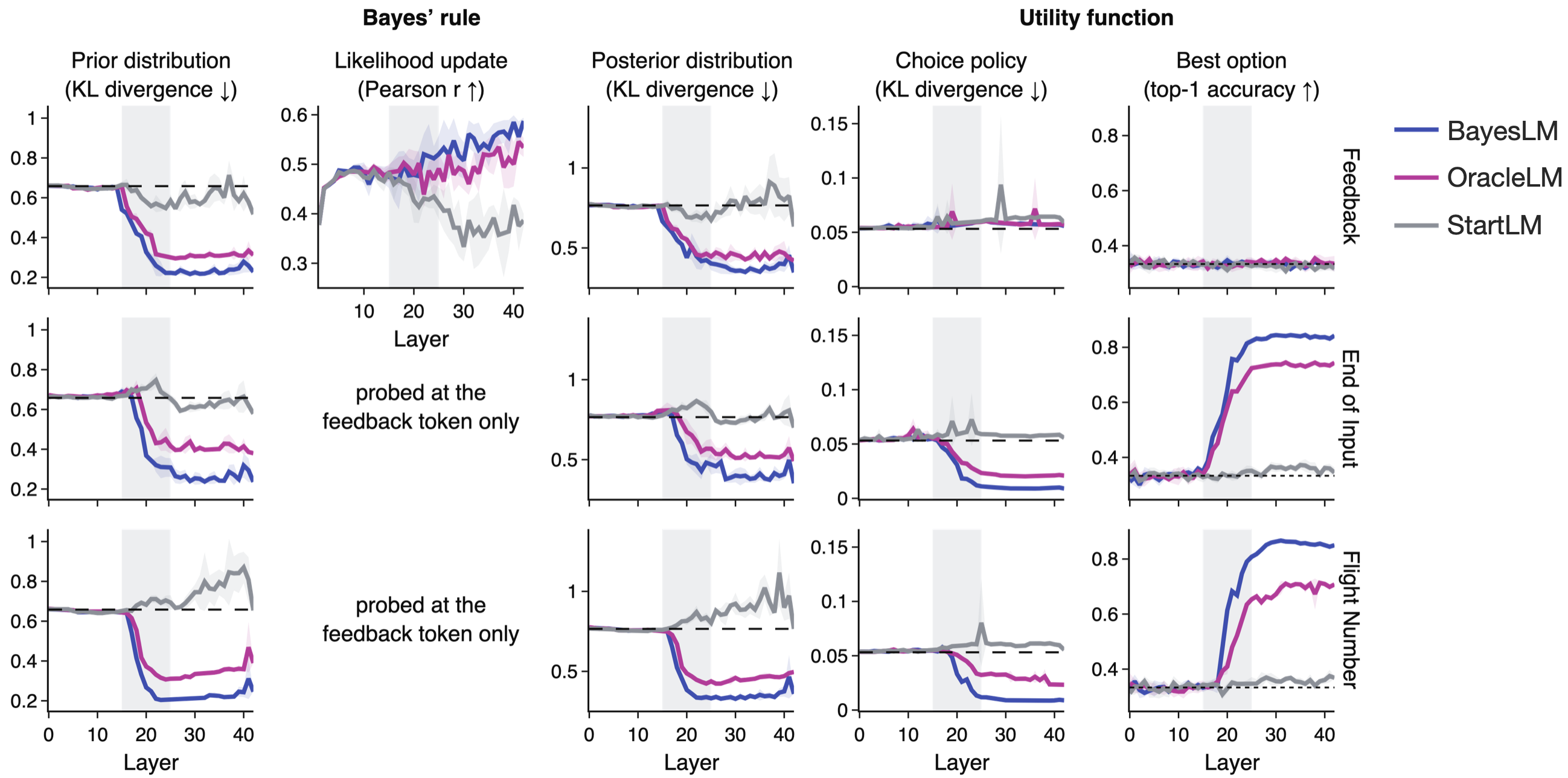}
  \caption{\textbf{Per-layer probing across token positions (detail for
  \autoref{fig:probing-overview}).} Each row is a read-out position (feedback token,
  last prompt token, digit-generation token); columns first present probed quantities of Bayes' rule, and then the probed policy and resulting prediction (argmax of probed policy).
  The choice representation (flight-choice KL, top-1) develops in the mid-to-late layers at the last-prompt and digit tokens --- where the \bayeslm{} and \oraclelm{} models rise well above \baselm{} --- but stays near chance at the feedback token, showing the choice is not yet committed there.
  Belief targets (prior/posterior KL) are decodable across all positions and depths. }
  \label{fig:detailed-token}
\end{figure*}

\begin{figure}[t]\centering
  \includegraphics[width=\linewidth]{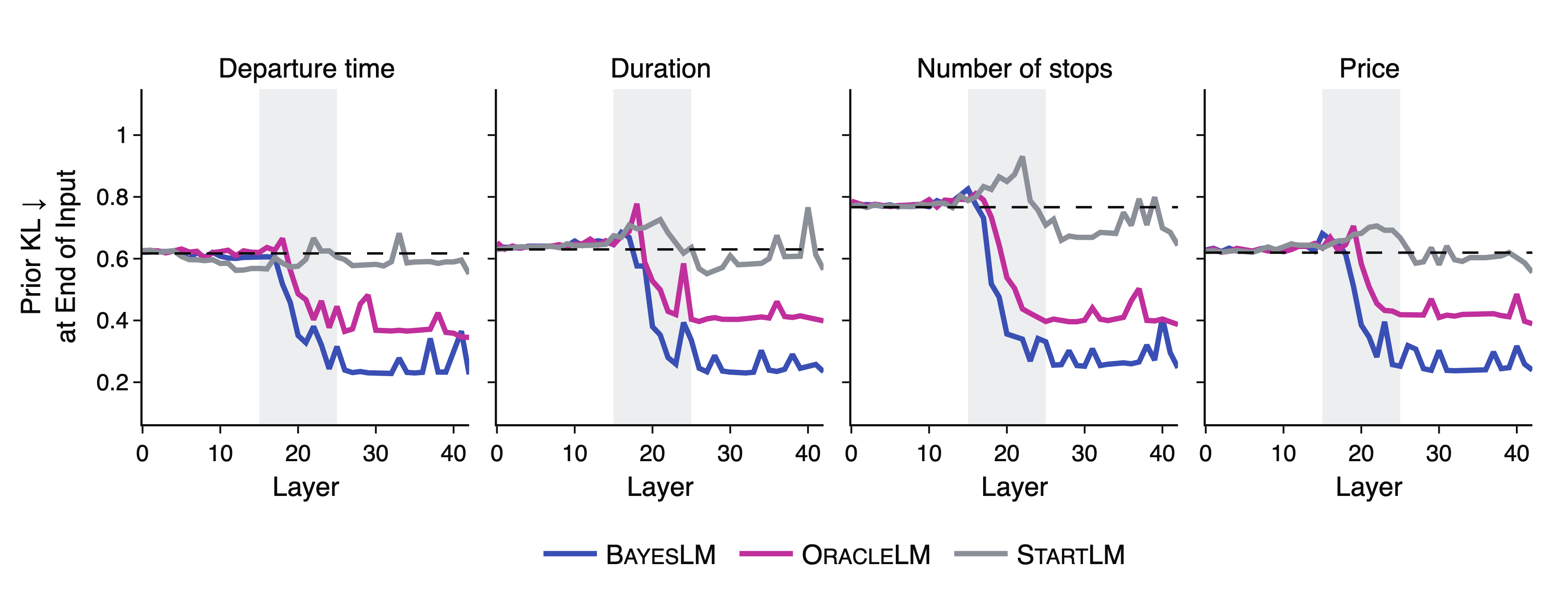}
  \caption{\textbf{Prior probing per feature.} KL of the prior probe per layer at the last-input token, on the original set. Dashed: a uniform probe for that feature; grey band: layers 15--25.}
  \label{fig:app-prior-features}
\end{figure}

\paragraph{Probing on additional test sets.} 
The main results in \autoref{sec:representing-beliefs} are only for specific token positions. 
\autoref{fig:detailed-token} shows probing results on the original set broken down by different probing position.
The choice representations (policy divergence, top-1 choice) develop in the mid-to-late layers at the last-prompt and digit tokens --- where the \bayeslm{} and \oraclelm{} models rise well above \baselm{} --- but stays near chance at the feedback token. This provides a sanity check that the choice is not yet committed before the new option set is seen.
Belief representations (prior/posterior divergence) are decodable across all positions and depths.

Additionally, we expect Bayes-consistent decoded beliefs to be order-invariant and for respond adequately to noisy input.
\autoref{fig:detailed-heldout} shows probing results broken down by test set. 
We indeed find that reordering the evidence leaves the belief divergence, averaged over layers, unchanged for \bayeslm{} and \oraclelm{}: 
a regression of belief divergence against the main effect of test set shows no significant effects ($p=0.49$ and $p=0.43$).
Under low- or high-noise, the decoded beliefs of both \bayeslm{} and \oraclelm{} have lower entropy than \bayesianassistant{}'s, resulting in a high divergence (for \bayeslm{}: $0.52$ on low-noise, and $0.58$ on high-noise.)
However, with exponential scaling of the probed beliefs with $\beta \approx 0.2$, \bayeslm{}'s belief divergence is $0.083$ (on low-noise) and $0.059$ (on high-noise). 
In parallel with \autoref{sec:acting-bayesian}, \bayeslm{} and \oraclelm{} mainly differ in how strongly they encode beliefs, but both are invariant to evidence order and noise and thereby Bayes-consistent. 
We speculate that the low default entropy of the probed beliefs might be an artifact of training probes on noise-free data. 
The \oraclelm{} behaves the same way.

\autoref{sec:representing-beliefs} averages the prior probe over the four flight features. \autoref{fig:app-prior-features} shows each feature separately.
The pattern is the same for all four.
Up to layer 15, no model does better than a uniform probe.
Between layers 17 and 25, the probe loss of the \bayeslm{} and the \oraclelm{} drops well below the uniform probe and then stays flat until the last layer.
The \bayeslm{} ends lowest on every feature (KL $\approx 0.25$), and the \oraclelm{} ends at about $0.40$.
The \baselm{} stays close to the uniform probe throughout, and in the belief window it even goes above it.
The features differ only in how high the uniform probe sits: the targets for the number of stops are sharpest ($0.77$, against $0.62$--$0.63$ for the other three). The fine-tuned models still reach the same final loss there, so they gain the most on this feature.
So the averaged result in the main text is not driven by any single feature.

\section{Additional Results on Belief Intervention}
\label{app:belief-intervention}

This section provides details on \autoref{sec:acting-on-beliefs}.
We describe the setup, report the hyperparameter search, compare prototype patching with additive steering, and list the results per feature and direction.

\paragraph{Setup.}
For each feature, direction, and layer, we build the prototype $v$ as the mean hidden state of the $500$ training instances with the highest prior on that feature (positive direction) or the lowest (negative direction).
We patch $v$ into the residual stream $h$ at the \textit{feedback} token and the next $32$ tokens across $8$ subsequent layers as $h' = (1 - \alpha)* h + \alpha*v$, using the same $v$ for all tokens and a separate one per layer.
We evaluate on all $624$ held-out instances.
Between $31\%$ and $38\%$ of them are instances where the \bayesianassistant{}'s recommendation changes under the patched belief, depending on the feature and direction.\footnote{We operationalize the edit in the \bayesianassistant{} by setting the $\text{belief}_4$ of the edited feature to 1 or -1, respectively.}
We treat this \bayesianassistant{}'s behavioral as the expected ground truth behavior under the intervention.
We measure the accuracy of the edited LMs' via \textit{flip accuracy}, which is the proportion of samples where the \bayesianassistant{} changes its recommendation upon intervention where the LM changes its recommendation upon intervention to the same option as the \bayesianassistant{}; and via \textit{stay accuracy}, which is the proportion of samples where the \bayesianassistant{} does not change its recommendation upon recommendation for which the LM does not change the recommendation, too. 
We report \textit{balanced accuracy}, which is their mean.
A baseline that never changes its recommendations scores $0.5$.

\begin{figure}[t]
  \centering
  \includegraphics[width=0.6\linewidth]{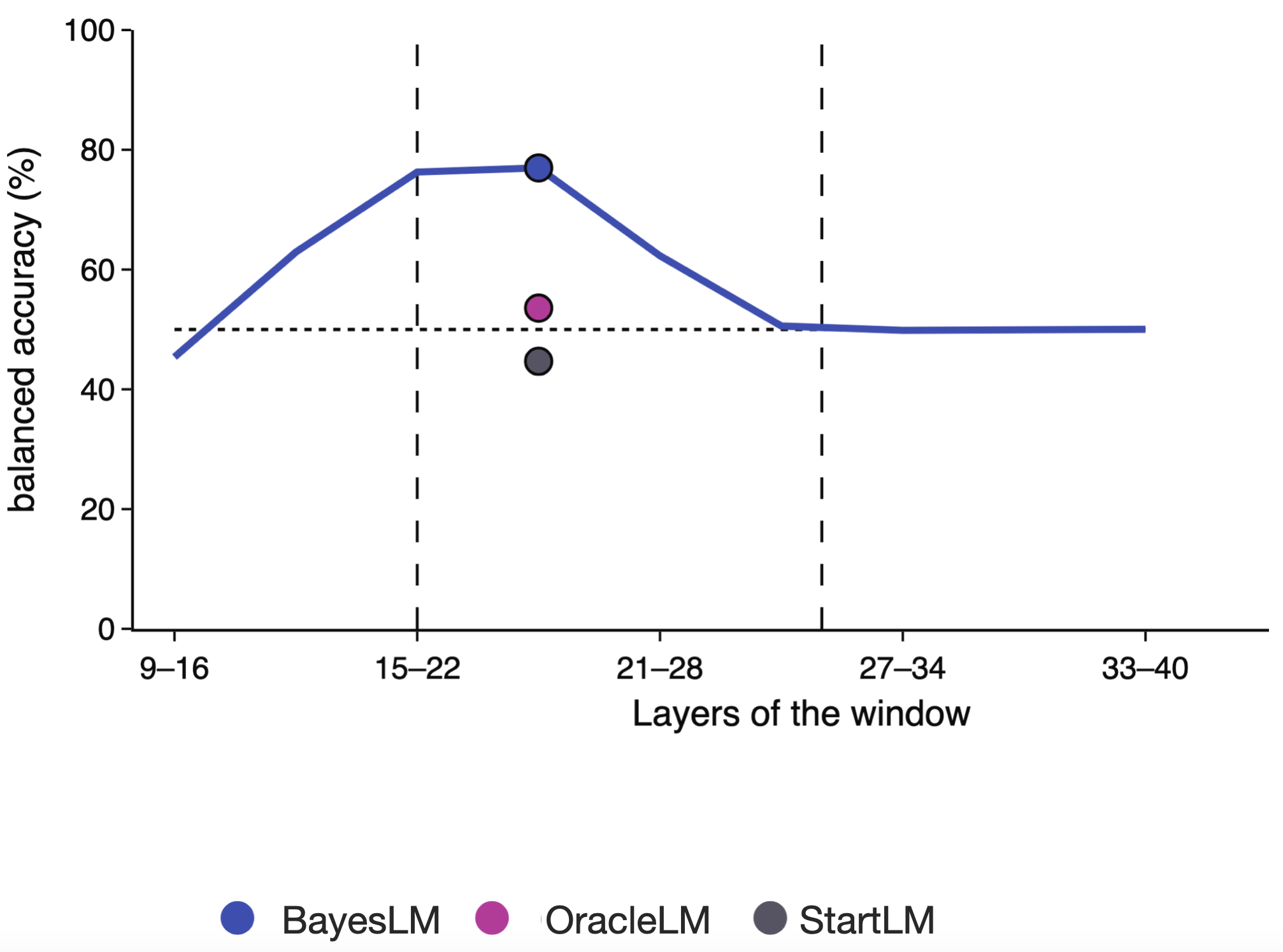}
  \caption{\textbf{Intervening on the encoded belief changes the recommendation in \bayeslm{}.} Balanced accuracy results of prototype belief patching. 
  }
  \label{fig:patching-results}
\end{figure}

\begin{figure}[t]
  \centering
  \includegraphics[width=0.7\linewidth]{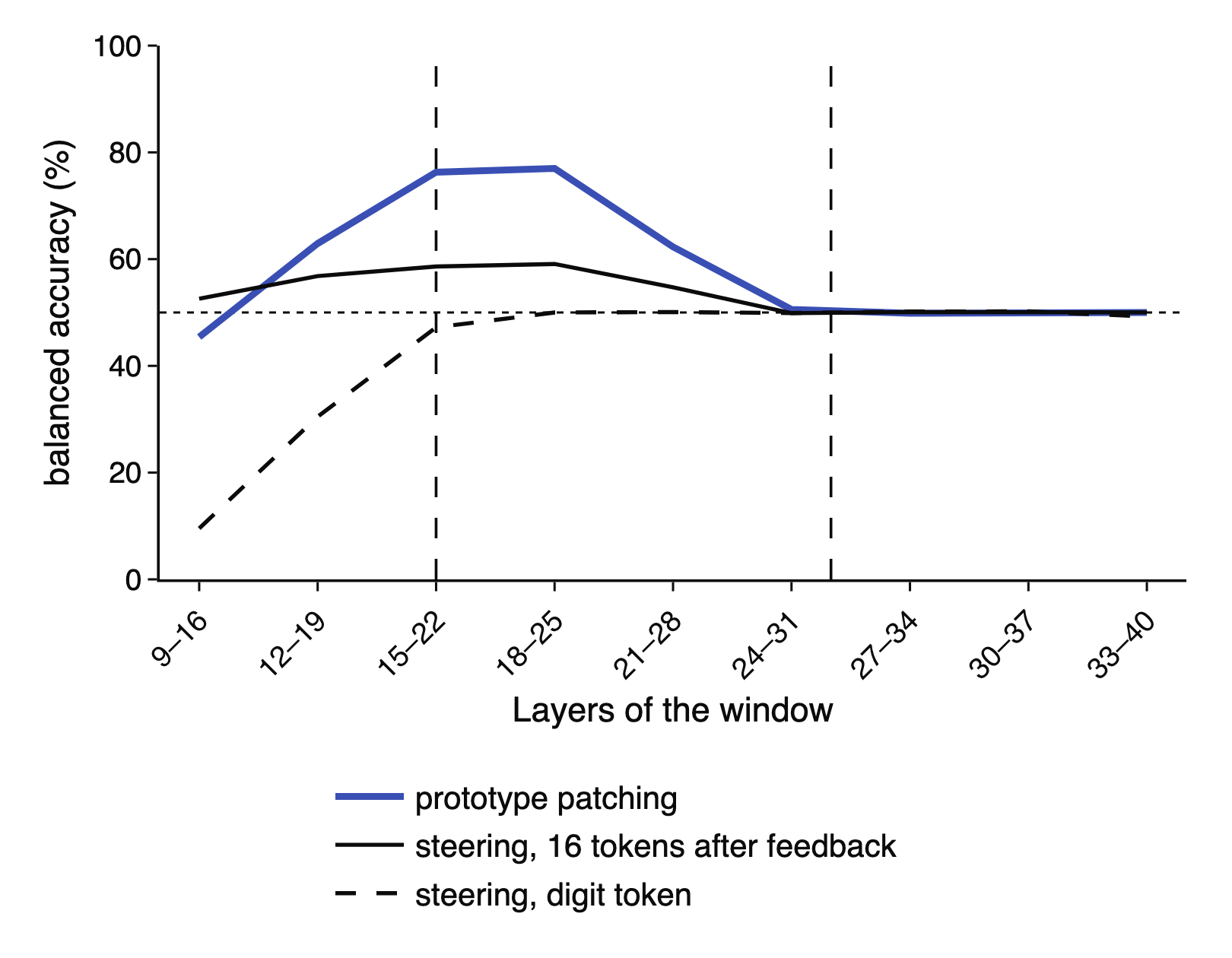}
  \caption{\textbf{Prototype patching versus CAA steering.}
  Balanced accuracy of the \bayeslm{} by layer window on the original test set, averaged over the eight feature-direction settings.
  Prototype patching uses $32$ tokens and $\alpha = 0.5$; additive steering uses $\alpha = 0.25$ at the $16$ tokens after the \textit{feedback} token or at the digit token.
  The dotted line marks a model that never changes recommendations; the dashed lines mark the layers where \autoref{sec:representing-beliefs} decodes the belief.}
  \label{fig:steering-vs-patching}
\end{figure}

\paragraph{Hyperparameter search.}
We tune the intervention parameters on the duration feature in the positive direction, using a random draw of $100$ instances where the recommendation should change and $100$ where it should remain the same. The parameters and results are shown in \autoref{tab:patching-hparams}.
We vary the dose $\alpha$ which weighs the injected vector $v$, the number of patched layers (window width), and the number of patched tokens after the \textit{feedback} token.
As shown in \autoref{tab:patching-hparams}, patching a single layer or only $4$ tokens changes few recommendations, so the belief has to be replaced across tokens and layers.
Larger doses raise flip accuracy but lower stay accuracy.
Therefore, we select $32$ tokens, $\alpha = 0.5$ and $8$ layers. The main results with these setting are visualized in \autoref{fig:patching-results}. 
\begin{table}[t]
\centering
\small
\begin{tabular}{lrrrr}
\toprule
setting & flip accuracy & stay accuracy & balanced accuracy  & noise effect \\
\midrule
\multicolumn{5}{l}{\emph{Varied dose $\alpha$, at 32 tokens and 8 layers}} \\
$\alpha = 0.25$ & 0.439 & 0.953 & 0.696 & 0.037 \\
\rowcolor{gray!12} \textbf{$\alpha = 0.50$} & 0.817 & 0.847 & 0.832 & 0.037 \\
$\alpha = 0.75$ & 0.939 & 0.800 & 0.870 & 0.037 \\
$\alpha = 1.00$ & 0.963 & 0.765 & 0.864 & 0.037 \\
\addlinespace
\multicolumn{5}{l}{\emph{Varied dose $\alpha$, at 4 tokens and 8 layers}} \\
$\alpha = 0.25$ & 0.098 & 0.988 & 0.543 & 0.012 \\
$\alpha = 0.50$ & 0.244 & 0.965 & 0.604 & 0.012 \\
$\alpha = 0.75$ & 0.280 & 0.976 & 0.628 & 0.012 \\
$\alpha = 1.00$ & 0.280 & 0.976 & 0.628 & 0.012 \\
\addlinespace
\multicolumn{5}{l}{\emph{Varied window width, at 32 tokens and $\alpha = 1$}} \\
1 layer  & 0.598 & 0.894 & 0.746 & 0.000 \\
2 layers & 0.683 & 0.871 & 0.777 & 0.012 \\
4 layers & 0.817 & 0.847 & 0.832 & 0.012 \\
8 layers & 0.963 & 0.765 & 0.864 & 0.037 \\
\addlinespace
\multicolumn{5}{l}{\emph{Varied window width, at 16 tokens and $\alpha = 1$}} \\
1 layer  & 0.439 & 0.906 & 0.672 & 0.000 \\
2 layers & 0.463 & 0.918 & 0.691 & 0.012 \\
4 layers & 0.683 & 0.918 & 0.800 & 0.012 \\
8 layers & 0.780 & 0.871 & 0.826 & 0.012 \\
\addlinespace
\multicolumn{5}{l}{\emph{Varied window width, at 4 tokens and $\alpha = 1$}} \\
1 layer  & 0.159 & 0.941 & 0.550 & 0.000 \\
2 layers & 0.195 & 0.941 & 0.568 & 0.012 \\
4 layers & 0.317 & 0.953 & 0.635 & 0.012 \\
8 layers & 0.280 & 0.976 & 0.628 & 0.012 \\
\bottomrule
\end{tabular}
\caption{\textbf{Hyperparameter search for prototype patching.}
We select the best hyperparameter for patching using an iterative search, varying one parameter at a time while keeping the rest fixed. 
We used a separate set of $100$ instances where the recommendation should change and $100$ where it should remain the same, generated by the \bayesianassistant{}.
We report for every setting, accuracy on instances that should flip and those that should stay, as well as the effect of injecting a norm-matched random vector (noise effect),
We use $32$ tokens, $\alpha = 0.5$ and $8$ layers. 
Based on these results, we choose to patch with $\alpha = 0.5$ on $32$ tokens across $8$ layers, which offers the best trade-off.}
\label{tab:patching-hparams}
\end{table}
\paragraph{Comparison with CAA steering.}
We also test contrastive activation addition steering \citep{rimsky2024steering}, which constructs a direction $d$ as the mean over differences between LMs' activations for instances with high belief values and instances with low belief values for a particular feature. 
We use the same data for computing $d$ as in \autoref{sec:acting-on-beliefs}.
$d$ is then added as $h' = h + \alpha*d$.
We use $\alpha = 0.25$ and the same $8$-layer windows, and inject either at the $16$ tokens after the \textit{feedback} token or at the digit token where the recommendation is generated.
Additive steering after the feedback token reaches at most $0.59$ balanced accuracy, against $0.77$ for prototype patching (\autoref{fig:steering-vs-patching}).
Steering at the digit token never exceeds chance, and in the early layers it corrupts the model away from the \bayesianassistant{}'s recommendation.
Both methods only work in the layers where the belief is decodable, but only prototype patching replaces the belief strongly enough to change most recommendations.

\begin{table}[t]
\centering
\small
\begin{tabular}{llrrr}
\toprule
 & & neuro-symbolic & $\Delta$ over LM  \\
\midrule
\multirow{2}{*}{Llama-3.1-8B}
 & \bayeslm{}  & 0.86 & $0.03$ \\
 & \oraclelm{} & 0.82 & $0.10$ \\
\addlinespace
\multirow{2}{*}{Qwen-2.5-7B}
 & \bayeslm{}  & 0.80 & $0.07$ \\
 & \oraclelm{} & 0.74 & $0.06$ \\
\bottomrule
\end{tabular}
\caption{\textbf{Neuro-symbolic predictions.}
We read each model's prior beliefs from its hidden states with linear probes, turn the
four decoded distributions into per-attribute rewards, and pick the flight the
\bayesianassistant{}'s own rule would pick from them. \textit{neuro-symbolic} is how often
that flight matches the \bayesianassistant{}'s recommendation, at the layer where each
model scores best. \textit{$\Delta$ over LM} is the gain over the same model answering on its
own.}
\label{tab:nesy-llama-qwen}
\end{table}

\paragraph{Results per feature and direction.}
Every result in \autoref{sec:acting-on-beliefs} averages the eight feature-direction settings, which \autoref{tab:patching-per-cell} lists individually.
The \bayeslm{} achieves a balanced accuracy between $0.75$ and $0.87$ across all settings and test sets, so no single feature drives the average.
Patching the \oraclelm{} or patching the \bayeslm{}'s prior into another model stays well below those values in every setting.

\begin{table}[t]
\centering
\small
\resizebox{\linewidth}{!}{%
\begin{tabular}{llrrrrrrrr}
\toprule
& & \multicolumn{4}{c}{\bayeslm{}} & & & & \\
\cmidrule(lr){3-6}
feature & direction & original & shuffled & low noise & high noise & \oraclelm{} & \bayeslm{}$\to$\oraclelm{} & \bayeslm{}$\to$\baselm{} & steering \\
\midrule
costs & negative & 0.775 & 0.799 & 0.817 & 0.818 & 0.566 & 0.582 & -- & 0.549 \\
costs & positive & 0.769 & 0.765 & 0.766 & 0.766 & 0.530 & 0.612 & -- & 0.589 \\
departure time & negative & 0.760 & 0.763 & 0.751 & 0.782 & 0.558 & 0.662 & -- & 0.565 \\
departure time & positive & 0.756 & 0.773 & 0.787 & 0.773 & 0.533 & 0.590 & -- & 0.628 \\
duration & negative & 0.766 & 0.770 & 0.785 & 0.786 & 0.505 & 0.606 & 0.525 & 0.613 \\
duration & positive & 0.759 & 0.761 & 0.786 & 0.809 & 0.530 & 0.604 & 0.584 & 0.597 \\
stops & negative & 0.768 & 0.777 & 0.829 & 0.841 & 0.493 & 0.593 & 0.439 & 0.595 \\
stops & positive & 0.806 & 0.817 & 0.841 & 0.866 & 0.570 & 0.649 & 0.535 & 0.590 \\
\midrule
mean & & 0.770 & 0.778 & 0.795 & 0.805 & 0.536 & 0.612 & 0.521 & 0.591 \\
\bottomrule
\end{tabular}}
\caption{\textbf{Balanced accuracy per feature and direction.}
Prototype patching at layers~$18$--$25$ with $32$ tokens and $\alpha = 0.5$.
The first four columns patch the \bayeslm{} with its own prior on each test set.
The next three use the original test set and patch the \oraclelm{} with its own prior, and the \bayeslm{}'s prior into the \oraclelm{} and the \baselm{}.
The last column is CAA steering at the $16$ tokens after the \textit{feedback} token ($\alpha = 0.25$) on the original test set.
The \baselm{} recipient was only run for duration and stops.}
\label{tab:patching-per-cell}
\end{table}

\paragraph{Interventions on various test sets.}
\autoref{tab:patching-per-cell} shows intervention results for each test set.
For \bayeslm{}, the intervention works across test sets: its balanced accuracy on the shuffled set is $0.778$, and $0.795$ and $0.805$ under low and high noise.
The noisy sets score slightly higher because \bayesianassistant{} changes its recommendation more often there, 
so more instances fall in the group where we expect a prediction change. 

\begin{figure*}[t]
  \centering
  \includegraphics[width=0.7\textwidth]{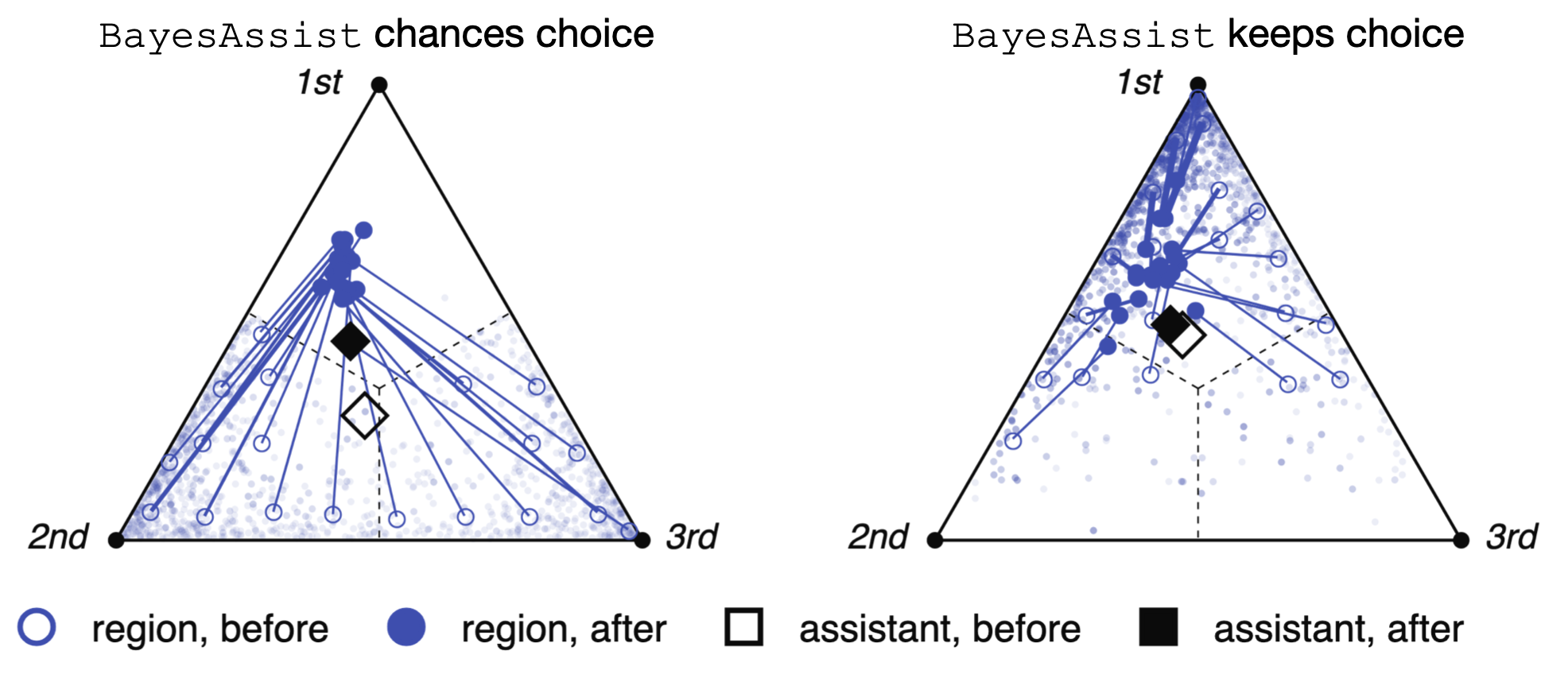}
  \caption{The choice policy on the simplex, with corners labeled by the \bayesianassistant{}'s ranking of the three options under the updated $\text{belief}_4'$, split by whether the \bayesianassistant{}'s recommendation changes after the intervention.
Open circles are the \bayeslm's mean policy before the intervention, filled circles after the intervention. 
The diamond marks the \bayesianassistant{}'s own policy before (open) and after (filled).}
  \label{fig:steering-overview}
\end{figure*}
\paragraph{Analyzing intervention failures in \bayeslm{}.}
We explore whether the success of the intervention depends on the magnitude of the required belief update. 
For this analysis, we focus on the instances where the \bayesianassistant{} changes its recommendation, and report the share on which the edited LM's prediction matches it. 
We group these instances into quartiles by the KL divergence between the original $\text{belief}_4$ and the patched $\text{belief}_4'$ in the \bayesianassistant{}.
The larger the divergence, the more often the LM moves: The LM changed its recommendation on $0.664$ of the instances in the smallest quartile (mean KL: $0.019$), and on $0.880$ of the instances in the largest (mean KL: $0.245$).
When the LM's recommendations change upon the intervention, they predict the correct option with equal frequency in each quartile (accuracy $0.83$--$0.85$).
So when the intervention fails, it does so because the model did not change its recommendation in the first place, not because it changed its prediction to the wrong option (see \autoref{fig:steering-overview}).

\section{Masking methods}
\label{app:masking}
In \autoref{sec:acting-on-beliefs}, we explore the computational route of Bayes' rule in the LMs.
To this end, we compare two mathematically equivalent calculations of the posterior given data $D_n$: the \textit{sequential} an the \textit{one-swoop} belief updating.
The \textit{sequential} update is presented in \autoref{eq:sequential-bayesian-assistant}, and accesses an intermediate beliefs after every step.
The \text{one-swoop} computation is shown in \autoref{eq:oneswoop} uses the whole $D_n$, without requiring intermediate belief representations:
\begin{equation}
P(\theta \mid D_n) \propto P_H(D_n \mid \theta) \times P(\theta) = \prod_{i=1}^{n} P_H(c_i \mid o_i, \theta) \times P(\theta)
\label{eq:oneswoop}
\end{equation}
While mathematically equivalent, in practice, using one or the other computation might not be innocuous: in an LM, a sequential computation could plausibly entail that the current beliefs are represented and used at specific context positions, whereas we assume that a one-swoop computation sources information across all tokens in the context.
We investigate whether the Bayes LM processes information in a way that is consistent with sequential or one-swoop posterior computation by restricting the information flow in the model's processing of the input through attention masking.
We expect that attention masking that is \textit{less} consistent with the model's computations will lead to a \textit{bigger} drop in the Bayes accuracy of its predictions.

\begin{figure*}[h]
    \centering
    \includegraphics[width=\linewidth]{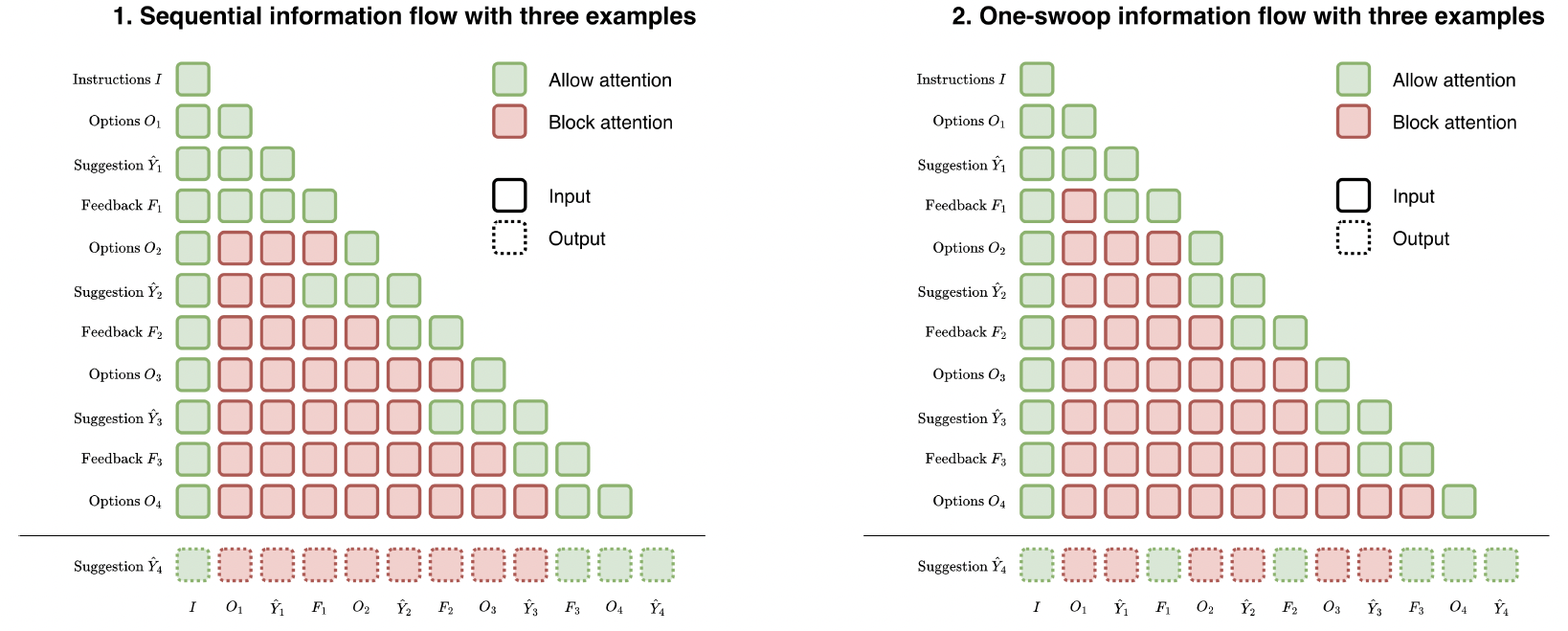}
    \caption{Overview of the attention masks applied for investigating how posterior beliefs might be formed.}
    \label{fig:attention-masks}
\end{figure*}

Intuitively, the sequential computation posits that posterior beliefs are calculated as soon as evidence is encountered, and access to these beliefs is only needed for making a prediction on a next sample. 
That is, attending to option sets and feedback for them on turn $i$ can be disabled after turn $i+1$.
One-swoop calculation, on the other hand, requires only accessing the likelihood of all observations at the point where the calculation of the belief for a new prediction would occur, allowing to block attention between a recommendation and the feedback at intermediate conversation rounds preceding the target prediction.
The masks operationalizing these restrictions are shown in \autoref{fig:attention-masks}.
The key manipulation is which tokens can access the feedback (i.e., evidence) tokens $F_i$: the sequential mask allows each next recommendation to attend only to the previous feedback, while the one-swoop mask allows attending to all previous options with their feedback for making the prediction at round five. 

\begin{figure*}[h]
    \centering
    \includegraphics[width=\linewidth]{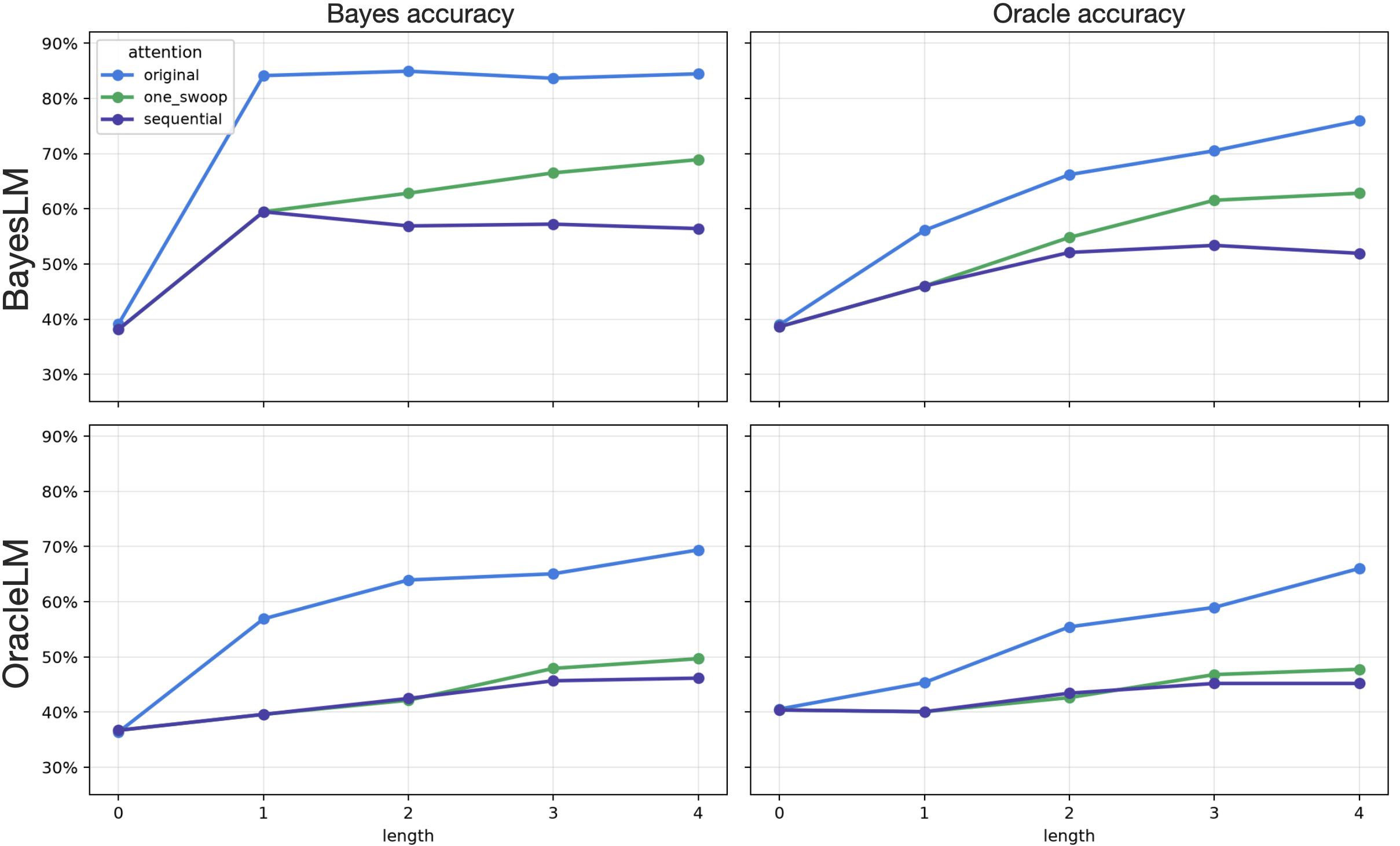}
    \caption{Overview of the attention masking accuracies (y-axis) of the \bayeslm{} and \oraclelm{}, across conversational rounds (x-axis).}
    \label{fig:attention-masks-results}
\end{figure*}

\begin{figure*}[h]
    \centering
    \includegraphics[width=\linewidth]{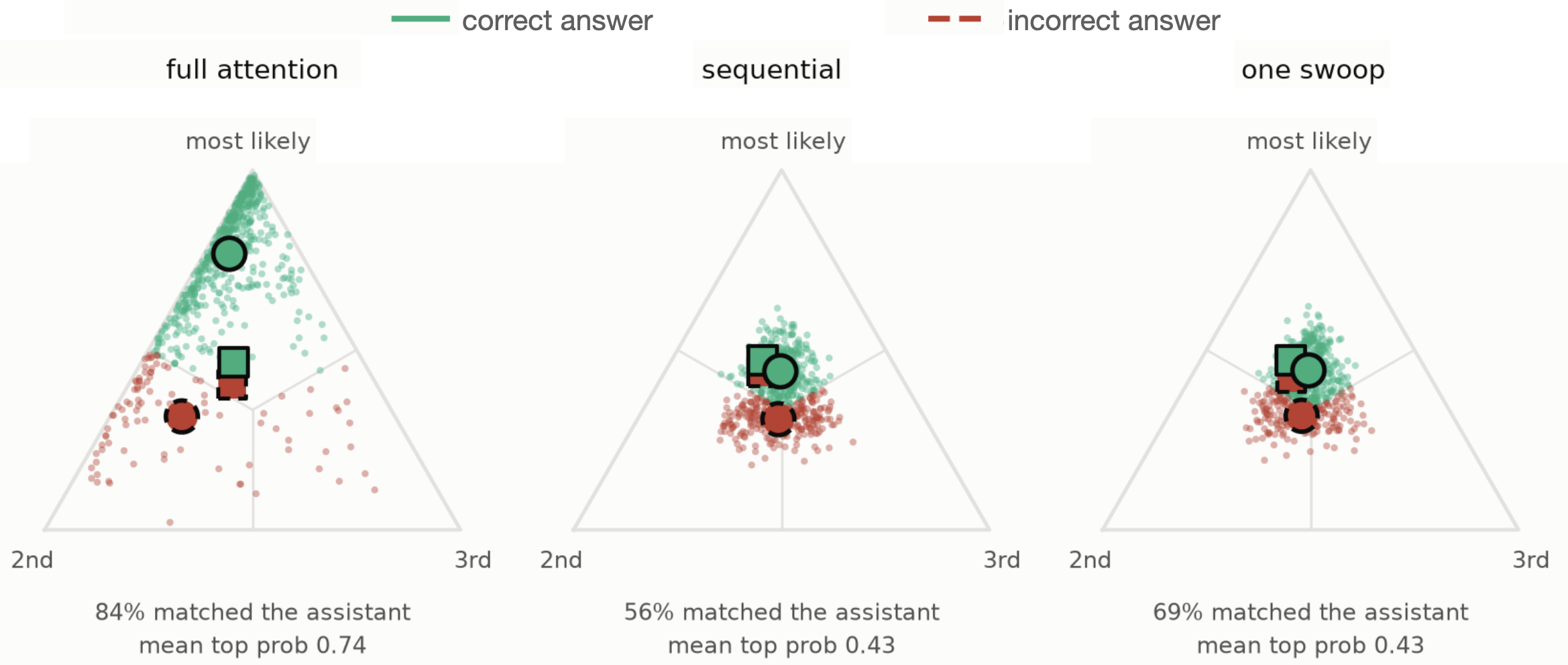}
    \caption{Distribution of the choice policy of the \bayeslm{} under masking.}
    \label{fig:attention-masks-simplex}
\end{figure*}

Bayesian fine-tuning seems to enable one-swoop, rather than sequential, forming of posterior beliefs: we observe a drop in Bayes accuracy of 16\% when the one-swoop mask is applied to the Bayes LM, whereas the drop is 28\% when the sequential mask is applied (see Figure~\ref{fig:attention-masks-results}).
The behavioral choice policies resulting from the masks are shown in \autoref{fig:attention-masks-simplex}.
Randomly blocking the same number of tokens causes a drop to an accuracy of 28-38\% (depending on the conversation length where it is applied), confirming that both of these masks focus on relevant information flow.



\section{Neuro-symbolic behavioral policy}
\label{app:nesy}
\begin{figure*}[t]
  \centering
    \includegraphics[width=\linewidth]{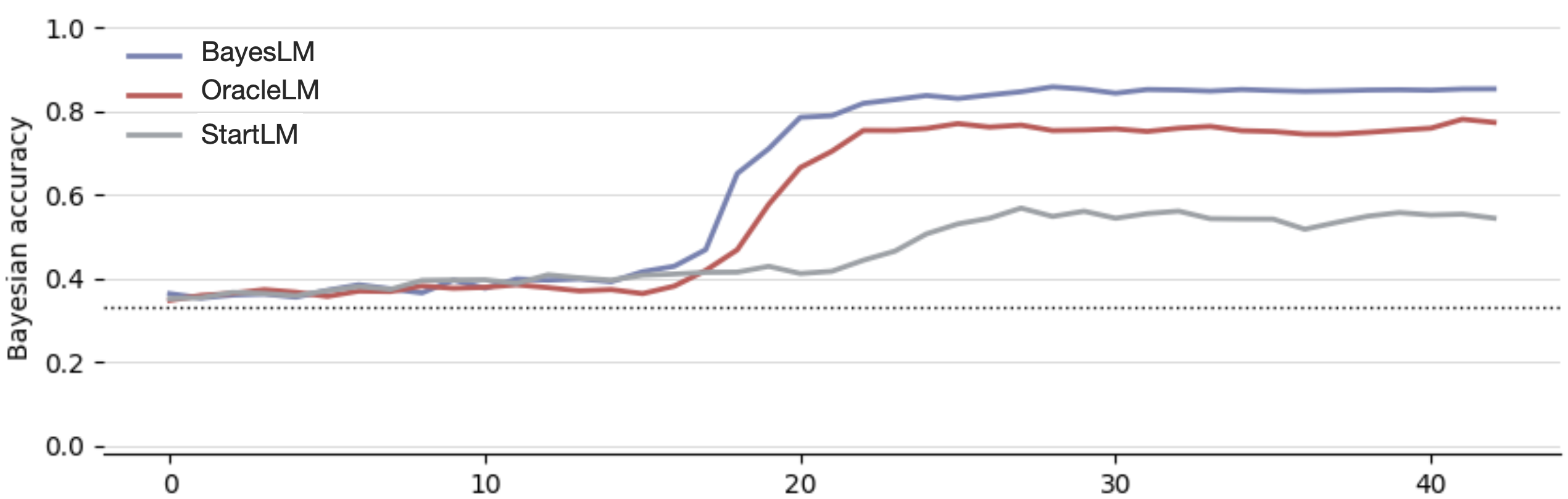}
  \caption{\textbf{Results of neuro-symbolic predictions across layers.} The development of the average Bayes accuracy of the three neuro-symbolic models when the probes stem from different model layers.}
  \label{fig:nesy-results}
\end{figure*}

The neuro-symbolic policies reported in \autoref{sec:bayesian-implementation} were calculated using probed beliefs from the last layer of Gemma. 
We additionally calculate Bayes accuracy of neuro-symbolic policies based on beliefs from different layers and show them in \autoref{fig:nesy-results}. This confirms that a neuro-symbolic policy can be derived once the beliefs are decodable, from about layer 15 on.

\section{Results across backbone models}\label{app:backbones}

\begin{figure}[h]
    \centering
    \includegraphics[width=\textwidth]{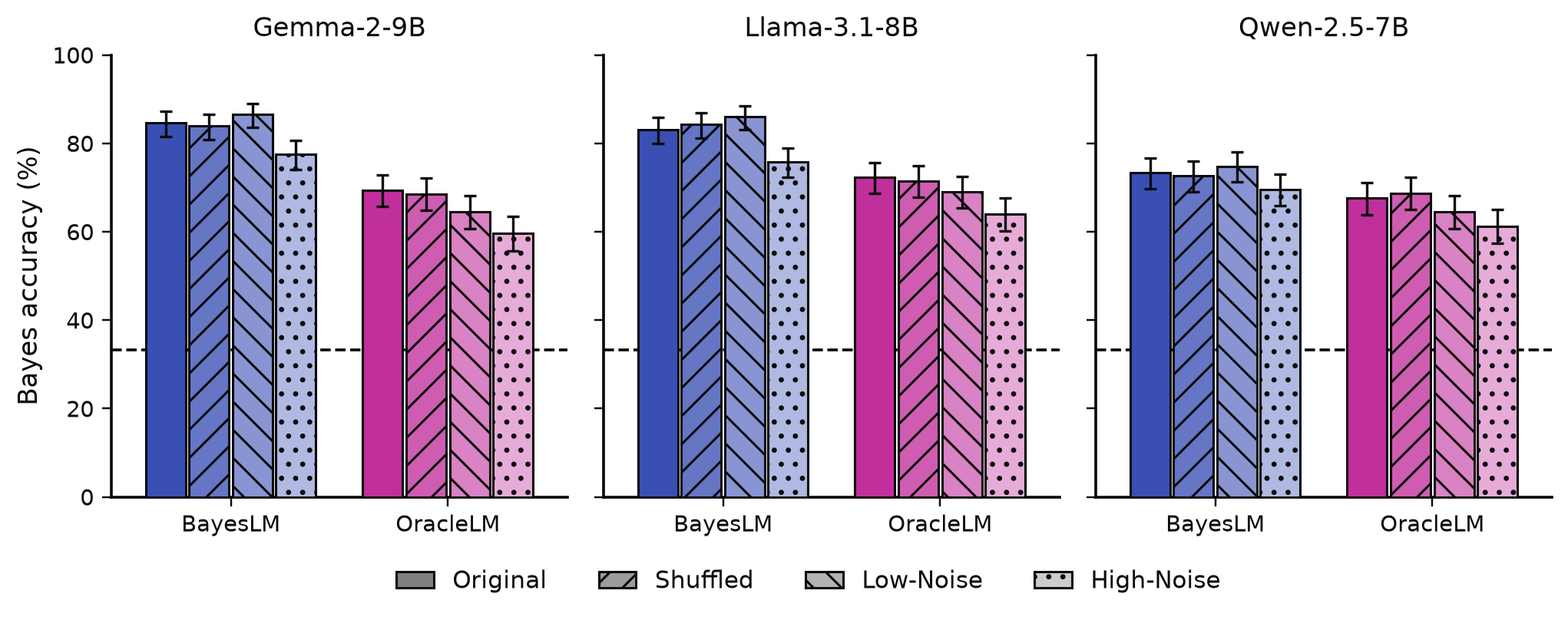}
    \caption{\textbf{Bayes accuracy across backbones (R1, \autoref{sec:acting-bayesian}).}
    Bayes accuracy of the \bayeslm{} and \oraclelm{} on the four test sets, with 95\% confidence intervals.
    Chance is $1/3$ (dashed line).
    The \bayeslm{} is more Bayes-accurate than the \oraclelm{} on every backbone, and is invariant to evidence order, and behaves as expected under noise.}
    \label{fig:app-bayes-accuracy-backbones}
\end{figure}

\begin{figure}[h]
    \centering
    \includegraphics[width=\textwidth]{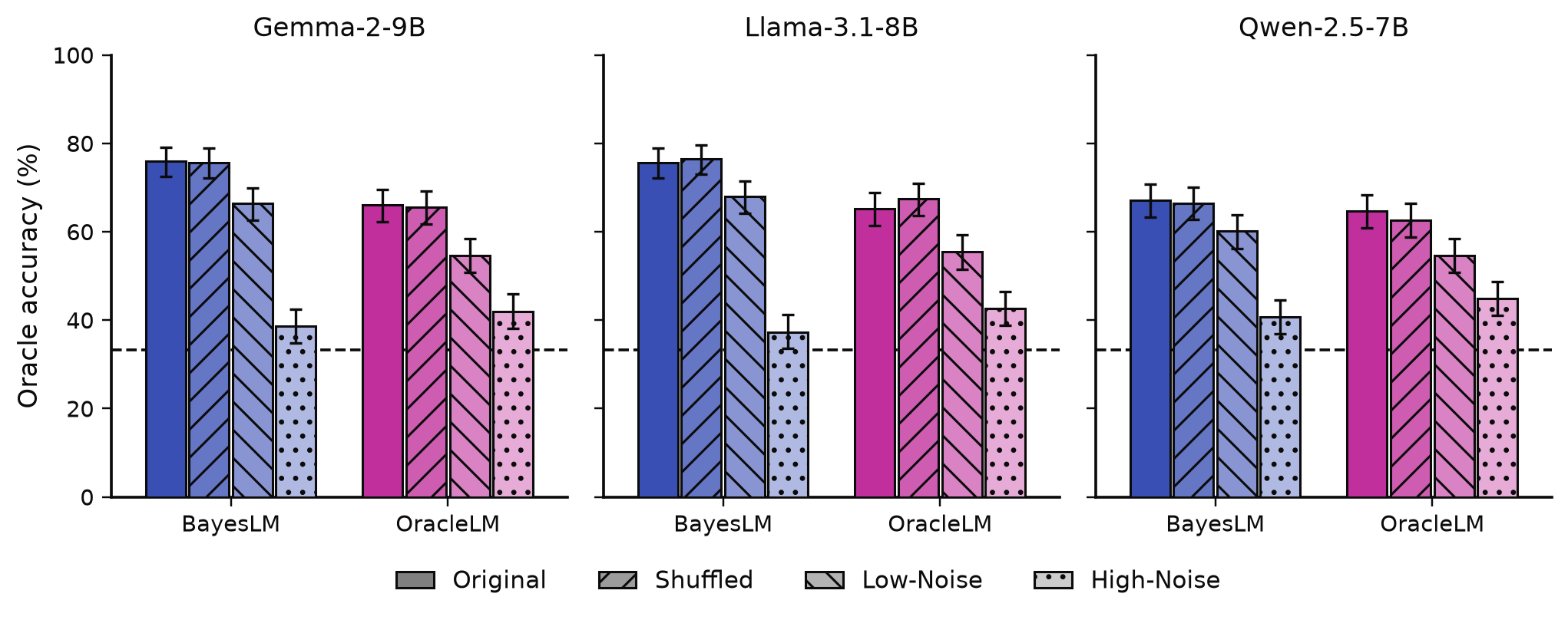}
    \caption{\textbf{Oracle accuracy across backbones (R1, \autoref{sec:acting-bayesian}).}
    Oracle accuracy shows the proportion of recommendations matching the user's true preference.
    All models lose accuracy in the presence of noise, as the \bayesianassistant{} does, because the evidence is less informative of the user's preference.
    The \bayeslm{} exceeds the \oraclelm{} on the original and shuffled sets for every backbone.}
    \label{fig:app-oracle-accuracy-backbones}
\end{figure}

\begin{figure}[h]
    \centering
    \includegraphics[width=\textwidth]{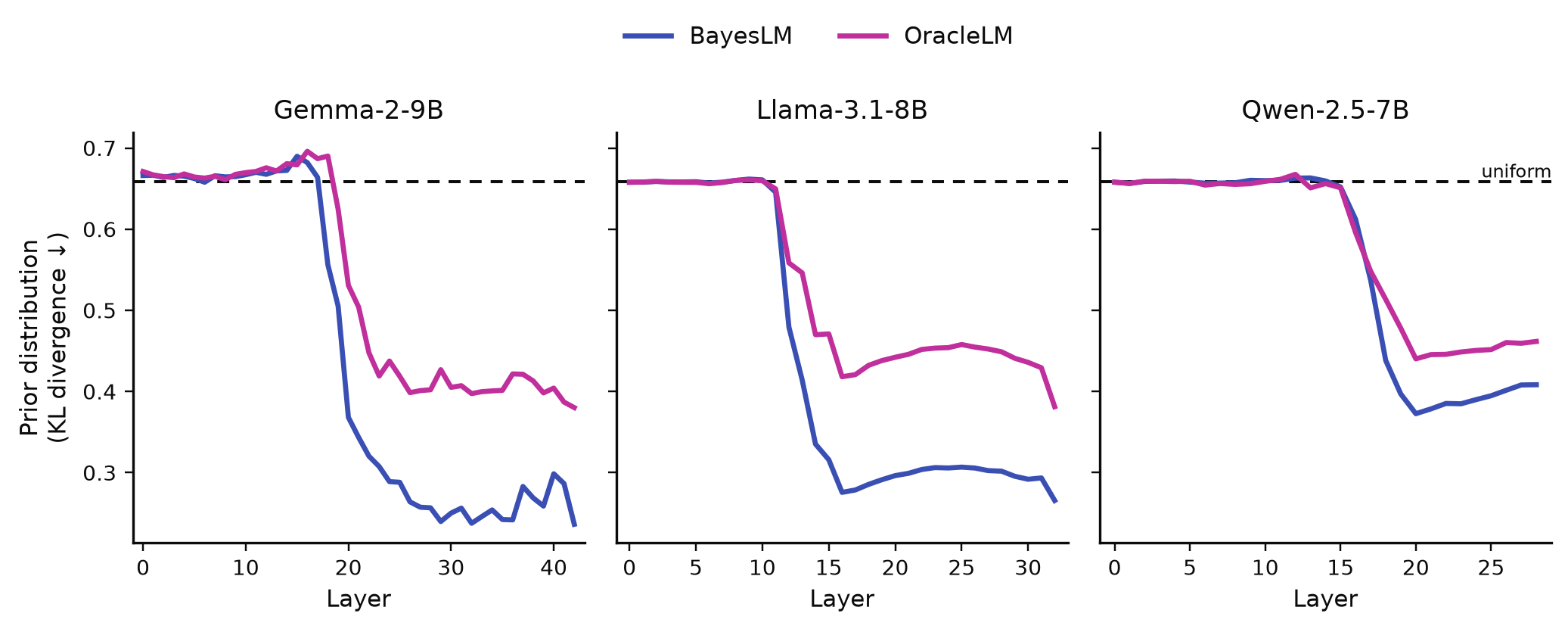}
    \caption{\textbf{Decoded prior across backbones (R2, \autoref{sec:representing-beliefs}).}
    Belief divergence between the decoded and the \bayesianassistant{}'s prior by layer, on the original set.
    The dashed line is the divergence of a uniform prior.
    On every backbone, the prior becomes decodable in the middle layers and more strongly for the \bayeslm{} than for the \oraclelm{}.}
    \label{fig:app-prior-probing-backbones}
\end{figure}

\begin{figure}[h]
    \centering
    \includegraphics[width=\textwidth]{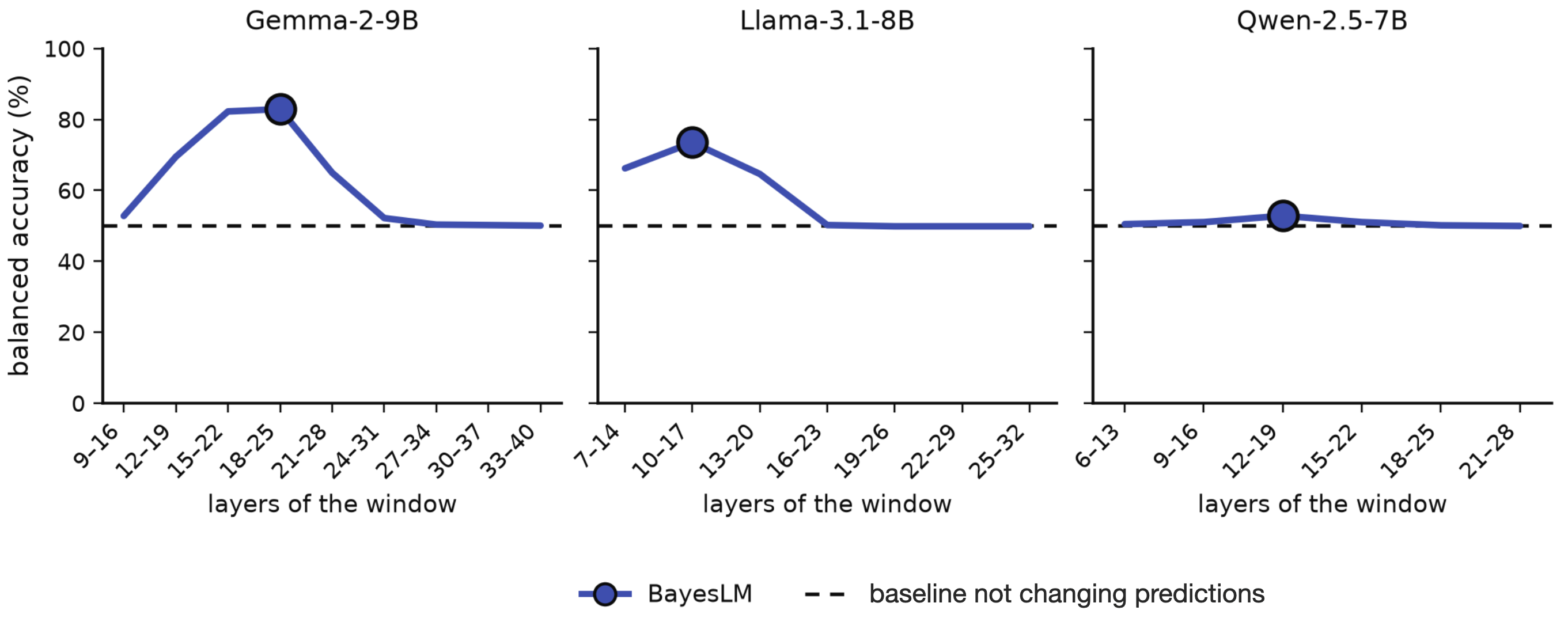}
    \caption{\textbf{Prior patching across backbones (R3, \autoref{sec:acting-on-beliefs}).}
    Balanced accuracy of the \bayeslm{} upon intervening on the prior, by the layer window of the edit.
    The dashed line is a model that never changes its recommendation.
    The marker shows the best window.
    The effect is largest for Gemma-2-9B, present for Llama-3.1-8B, and near chance for Qwen-2.5-7B.}
    \label{fig:app-steering-backbones}
\end{figure}

\begin{figure}[h]
    \centering
    \includegraphics[width=\textwidth]{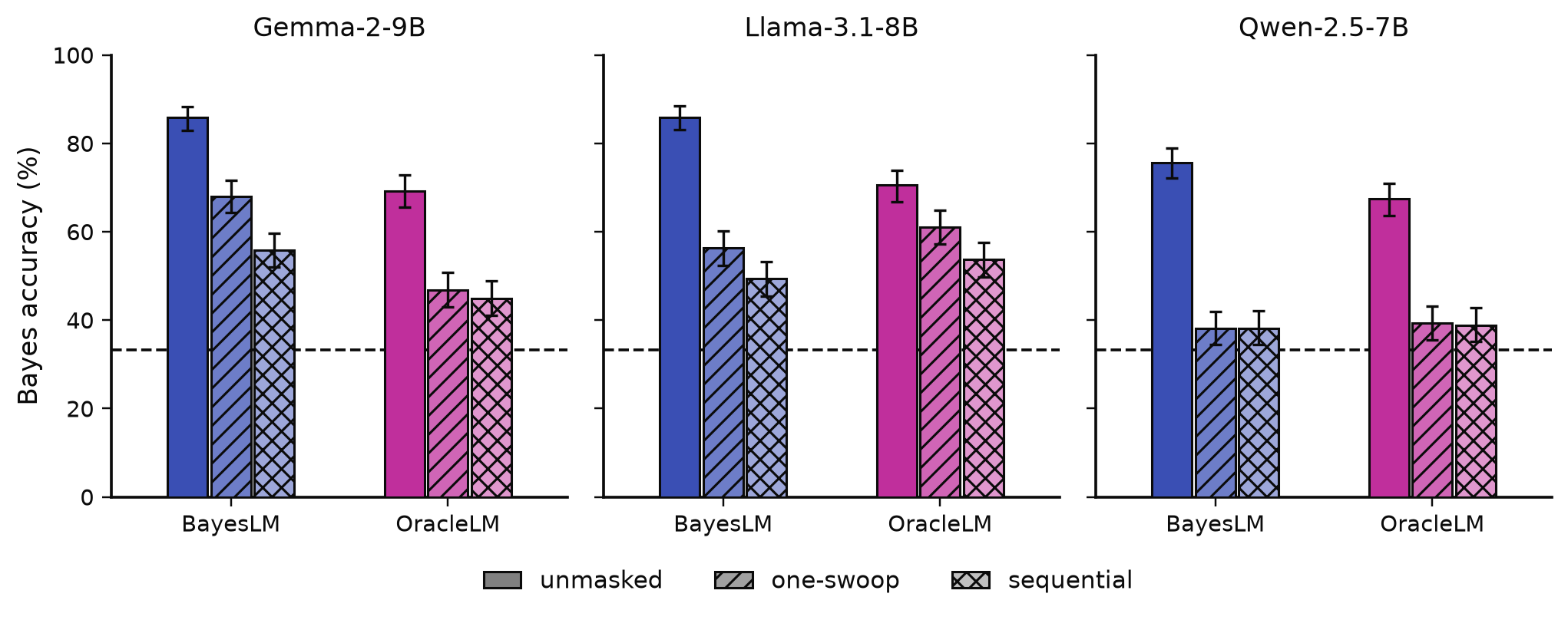}
    \caption{\textbf{Attention masking across backbones (R3, \autoref{sec:acting-on-beliefs}).}
    Bayes accuracy without masking and under the one-swoop and sequential masks.
    Chance is $1/3$ (dashed line).
    On every backbone, both masks reduce accuracy, and the sequential mask reduces it more, though the gap between the two masks is smaller for Llama-3.1-8B and Qwen-2.5-7B than for Gemma-2-9B.}
    \label{fig:app-masking-backbones}
\end{figure}

We repeat the main experiments addressing R1--R3 with Llama-3.1-8B-Instruct \citep{grattafiori2024llama} and Qwen-2.5-7B-Instruct \citep{qwen2025qwen25technicalreport} as the \baselm{}, each fine-tuned into a \bayeslm{} and an \oraclelm{} provided by \citet{qiu2026bayesian}.
All figures show Gemma-2-9B (which was reported in the main text) alongside the two other backbones.
Addressing R1 (acting Bayesian), \autoref{fig:app-bayes-accuracy-backbones} shows the LMs' Bayes accuracies on the different test sets, confirming that the \bayeslm{} is more Bayes-accurate than the \oraclelm{} on every backbone, and is invariant to evidence order, and behaves as expected under noise.
\autoref{fig:app-oracle-accuracy-backbones} shows the LMs' oracle accuracy across test sets.
The \bayeslm{} exceeds the \oraclelm{} on the original and shuffled sets for every backbone.
All models lose accuracy in the presence of noise, as predicted by the \bayesianassistant{}, because the evidence is less informative of the user's preference.
Addressing R2 (representing beliefs), \autoref{fig:app-prior-probing-backbones} shows the belief divergence between the decoded and the \bayesianassistant{}'s prior on the original test set.
On every backbone, the prior becomes decodable in the middle layers, and is encoded more strongly for the \bayeslm{} than for the \oraclelm{}.
Addressing R3 (acting on beliefs), \autoref{fig:app-steering-backbones} shows the balanced accuracy of the \bayeslm{} upon intervening on the prior.
The intervention effect is largest for Gemma-2-9B, present for Llama-3.1-8B, and near chance for Qwen-2.5-7B
\autoref{fig:app-masking-backbones} shows the LMs' Bayes accuracy without masking and under the one-swoop and sequential masks. On every backbone, both masks reduce accuracy, and the sequential mask reduces it more, though the gap between the two masks is smaller for Llama-3.1-8B and Qwen-2.5-7B than for Gemma-2-9B.
   
\end{document}